\pdfoutput=1

\documentclass[11pt]{article}

\usepackage{acl}

\usepackage{times}
\usepackage{tikz}  
\usepackage{latexsym}
\usepackage{booktabs}
\usepackage{graphicx}
\usepackage{xspace}
\usepackage{CJKutf8}
\usepackage{multirow}
\usepackage{multicol}
\usepackage{amsbsy}
\usepackage{amssymb}
\usepackage{amsmath}
\usepackage{pgfplots}
\usepackage{natbib}
\usepackage{verbatim}
\usepackage{pifont}
\usepackage[T1]{fontenc}

\usepackage[utf8]{inputenc}

\usepackage{microtype}
\usepackage[Symbol]{upgreek}
\usepackage{enumerate}
\newenvironment{itemize*}%
 {\leftmargini=10pt\begin{itemize}%
  \setlength{\itemsep}{0pt}%
  \setlength{\parskip}{0pt}%
  }%
 {\end{itemize}}
\newenvironment{enumerate*}%
 {\begin{enumerate}%
  \setlength{\itemsep}{0pt}%
  \setlength{\parskip}{0pt}}%
 {\end{enumerate}}

\usepackage{listings}

\newcommand{\toolname}{\textsc{FacTool}\xspace}

\definecolor{weizhey}{rgb}{0.43, 0.71, 0.40}

\title{\toolname : Factuality Detection in Generative AI \\ A Tool Augmented Framework for Multi-Task and Multi-Domain Scenarios}

\author{
I-Chun Chern\textsuperscript{\rm{2}} \quad Steffi Chern\textsuperscript{\rm{2}} \quad Shiqi Chen\textsuperscript{\rm{3}} \quad Weizhe Yuan\textsuperscript{\rm{4}} \quad Kehua Feng\textsuperscript{\rm{1}} \\ \textbf{Chunting Zhou}\textsuperscript{\rm{5}} \quad \textbf{Junxian He}\textsuperscript{\rm{6}} \quad \textbf{Graham Neubig}\textsuperscript{\rm{2}} \quad \textbf{Pengfei Liu}\textsuperscript{\rm{1,7}}\thanks{\ \ Corresponding author}\\
\textsuperscript{1}Shanghai Jiao Tong University \
\textsuperscript{2}Carnegie Mellon University \\ 
\textsuperscript{3}City University of Hong Kong \
\textsuperscript{4}New York University \
\textsuperscript{5}Meta AI \\
\textsuperscript{6}The Hong Kong University of Science and Technology \\
\textsuperscript{7}Shanghai Artificial Intelligence Laboratory \
}

\begin{document}
\maketitle
\begin{abstract}
The emergence of generative pre-trained models has facilitated the synthesis of high-quality text, but it has also posed challenges in identifying factual errors in the generated text. In particular: (1) A wider range of tasks now face an increasing risk of containing factual errors when handled by generative models. (2) Generated texts tend to be lengthy and lack a clearly defined granularity for individual facts. (3) There is a scarcity of explicit evidence available during the process of fact checking.

With the above challenges in mind, in this paper, we propose \toolname, a task and domain agnostic framework for detecting factual errors of texts generated by large language models (e.g., ChatGPT). Experiments on four different tasks (knowledge-based QA, code generation, mathematical reasoning, and scientific literature review) show the efficacy of the proposed method.
We release the code of \toolname associated with ChatGPT plugin interface at \url{https://github.com/GAIR-NLP/factool}.
\end{abstract}

\section{Introduction}
Generative artificial intelligence (AI) technology, exemplified by GPT-4 \cite{openai2023gpt4} consolidates various tasks in natural language processing into a single sequence generation problem. 
This unified architecture enables users to complete multiple tasks (e.g., question answering~\cite{thoppilan2022lamda}, code generation~\cite{chen2021evaluating}, math problem solving~\cite{lewkowycz2022solving}, and scientific literature generation~\cite{taylor2022galactica}) through a \textit{natural language interface}~\cite{liu2023pre} with both unprecedented performance~\cite{bubeck2023sparks} and interactivity. 

\begin{figure}[t]
    \centering
    \includegraphics[width=1.0\linewidth]{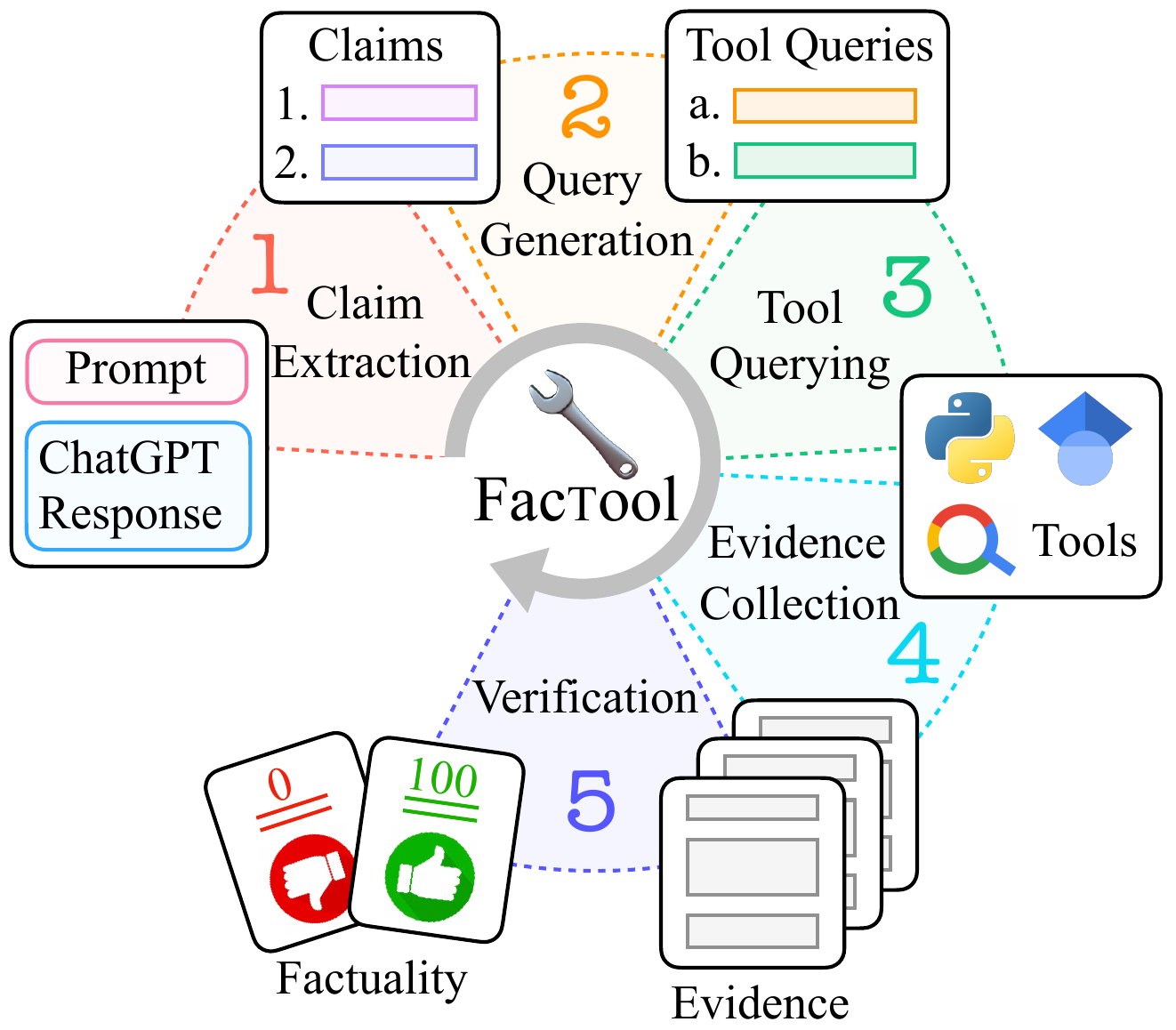}
    \caption{Tool-augmented framework for factuality detection.
    }
    \label{fig:intro}
\end{figure}

However, at the same time, such a \textit{generative paradigm} also introduces some unique challenges.
Content that is automatically generated can often exhibit 
inaccuracies or deviations from the truth
due to the limited capacity of large language models (LLMs)
\cite{ji2023survey, Schulman2023}. LLMs are susceptible to producing content that appears credible but may actually be factually incorrect or imprecise. 
This limitation restricts the application of generative AI in some high-stakes areas, such as healthcare, finance, and law.
Therefore, it is crucial to identify these errors systematically to improve the usefulness and reliability of the generated content.

\begin{table*}[!htbp]
  \centering
  \footnotesize
    \begin{tabular}{lcccccll}
    \toprule
    \multicolumn{1}{r}{\multirow{2}[4]{*}{\textbf{Methods}}} & \multicolumn{2}{c}{\textbf{Response}} & \multicolumn{2}{c}{\textbf{Claim}} & \textbf{Evidence} & \multicolumn{2}{c}{\textbf{Scenario}} \\
\cmidrule{2-3}  \cmidrule{4-8}       & \textbf{Length} & \textbf{Generated by} & \textbf{Granularity} & \textbf{Provided} & \textbf{Provided} & \multicolumn{1}{c}{\textbf{Domain}} & \textbf{Task} \\
    \midrule
    FEVER-based & 7.30 & Human & Fact   & \checkmark   & \sffamily X    & Wikipedia & Fact Verification \\
    FactCC & 20.83 & Synthetic & Sentence    & \checkmark    & \checkmark   & Newswire & Summ. Factuality \\
    QAGS-based & 16.11 & Model & Summary    & \checkmark    & \checkmark   & Newswire & Summ. Factuality \\
    WICE-based & 24.20 & Human & Fact   & \checkmark   & \checkmark    & Wikipedia & Entailment \\
    RARR  & - & PaLM/LaMDA  & Fact    & \sffamily X    & \sffamily X    & Wikipedia & QA \\
    \midrule
    \multirow{4}{*}{\toolname}  & 41.80 & ChatGPT & Fact   & \sffamily X    & \sffamily X    & Wikipedia & QA \\
      & 30.37  & ChatGPT & Snippet   &  \sffamily X   & \sffamily X    & Python & Code generation \\
     & 67.13  & ChatGPT & Statement   & \sffamily X    & \sffamily X    & Math & Math Problems \\
      & 76.34 & ChatGPT & Tuple   & \sffamily X    & \sffamily X    & Sci. text & Sci. Review \\
    \bottomrule
    \end{tabular}%
      \caption{A comparison of published approaches for factuality detection in terms of generated responses and claims to be verified based on collected evidence. ``Scenario'' represents which task and domain the corresponding approach has been justified.
      ``Sci.'' represents ``Scientific''.
      } 
  \label{tab:comparisons}%
\end{table*}%

Current literature on detecting and mitigating factual errors generated by machine learning models focuses predominantly on a single specific task, for example, retrieval-augmented verification models for QA \cite{lewis2020retrieval}, hallucination detection models for text summarization \cite{fabbri-etal-2022-qafacteval}, and execution-based evaluation for code \cite{shi-etal-2022-natural}.
While these methods have proven successful within their respective areas, given the remarkable \emph{versatility} of tasks and domains handled by LLMs, we argue that it is also important to have a more comprehensive factuality detection and verification framework that is similarly versatile.

Additionally, in the current literature, the task of factuality detection is usually simplified as either (i) given a claim, determining whether it is factually correct, (ii) or given evidence, determining whether the generated claim is supported. This task definition is not well suited to writing tasks that users commonly engage with when interacting with generative models (e.g., ChatGPT), where we often need to validate the factuality of a long-form generation \textit{without} explicit claims and evidence.

In this paper, we propose a task and domain-agnostic framework, \toolname, which aims to detect factual errors in LLM-generated texts. We illustrate our framework in Fig.~\ref{fig:intro}, where we connect the concept of ``\emph{tool use}''~\cite{thoppilan2022lamda,gao2022pal,schick2023toolformer} with ``\emph{factuality detection}'' and demonstrate that the ability to use tools in LLMs is crucial for factuality detection.
Specifically, \toolname leverages various tools, including Google Search, Google Scholar, code interpreters, Python, or even LLMs themselves, to gather evidence about the factuality of the generated content. Moreover, our framework employs the reasoning abilities of LLMs to assess the factuality of the content, given the evidence that has been gathered. We develop a benchmark and perform experiments across four tasks: knowledge-based QA, code generation, math problem solving, and scientific literature review writing. 

In summary, our contributions are:
\begin{itemize*}
    \item We revisit the task of factuality detection 
    and extend it in a way that allows for a better audit of current generative AI models.
    \item 
    We connect the concept of ``tool use'' with ``factuality detection'', developing a unified and versatile framework for factuality detection across a variety of domains and tasks.
    \item We use \toolname to evaluate the factuality of modern chatbots, and found that GPT-4 has the best factuality across almost all scenarios. Supervisely fine-tuned chatbots (Vicuna-13B) have reasonably good factuality in KB-based QA but perform poorly in more challenging scenarios, including code generation, math problem solving, and scientific literature review writing.
    
\end{itemize*}

\section{Related Work}

\paragraph{Factuality Detection in Natural Language Processing}

Factuality detection was a topic of rigorous study even before the advent of generative AI.  Existing works can be organized by their differences in terms of the ``response'' to be verified, the ``claim'' extracted from the response, and supporting ``evidence''.
As illustrated in Tab.~\ref{tab:comparisons}, the creation of the FEVER dataset~\cite{thorne-etal-2018-fever} spawned models~\cite{zhong-etal-2020-reasoning, krishna-etal-2022-proofver} that determine whether a given fine-grained claim made based on Wikipedia\footnote{\url{https://www.wikipedia.org/}} articles is correct. In this task setting, both the claim and related evidence are given.
FactCC~\cite{kryscinski-etal-2020-evaluating} and QAGS-based models ~\cite{wang-etal-2020-asking} adopted different task formulations to detect \textit{factual consistency}, i.e., given the evidence text, and the goal is to determine if the generated summaries or summary sentences are factually consistent with the given text.
WICE-based methods~\cite{kamoi2023wice} decide if a fact from a Wikipedia sentence could be supported by provided evidence.
RARR~\cite{gao2022rarr} proposed a new approach by directly prompting LLMs to generate queries, retrieve evidence and determine factuality.

Existing works typically rely on either a given claim or given evidence and target a specific use case. However, in this paper, we introduce a more challenging yet practical task setting, i.e., factuality detection without explicit claims or evidence, and propose a framework capable of addressing this challenge in a variety of scenarios.

\begin{table*}[!htbp]
  \centering
  \footnotesize
    \begin{tabular}{lllll}
    \toprule
    \textbf{Tasks} & \textbf{Prompt ($p$)} & \textbf{Response ($r$)} & \textbf{Claim ($c$)} & \textbf{Evidence ($e$)} \\
    \midrule
    KB-based QA & Question & Long-form answer & Atomic component unit & Web searched results \\
    Code Generation & Code Query & Executable code & Code snippet & Python library \\
    Math Problems & Math problems & Math solution & Math calculation & Calculator \\
    Sci. Lit Review & Scientific question & Long-form review & Tuple (paper title, year, authors) & Google scholar \\
    \bottomrule
    \end{tabular}%
      \caption{Factuality definition in different tasks. ``Sci. Lit Review'' represents scientific literature review. 
      }
  \label{tab:factuality-definition}%
\end{table*}%

\paragraph{Tool use in Large Pretrained Language Models}
Language models store limited knowledge within their parameters. To overcome this limitation, various tools have been introduced to assist language models in order to further expand their capabilities. For example, \citet{press2022measuring, komeili-etal-2022-internet} gathered information from the Internet to enhance question answering and dialog systems, respectively. \citet{schick2023toolformer} trained a model capable of interacting with five tools including a calculator, a translation system, etc. Recently, \citet{shen2023hugginggpt} introduced a framework that employs LLMs to connect various AI models from the machine learning communities to tackle AI tasks. Furthermore, \citet{liang2023taskmatrixai} proposed a new AI ecosystem that connects LLMs with millions of existing APIs to accomplish tasks. In this work, we explore tool use in LLMs for the task of factuality detection.

\section{Revisiting Factuality in Generative AI}

\subsection{Definition}

\paragraph{Versatile Factuality}
In most previous works, factuality has been defined as whether a claim in a text can be supported by evidence from a separate, trustworthy knowledge base, with applications in fact-checking~\cite{Thorne18Fever} (where the knowledge base is a large source like Wikipedia) and summarization~\cite{kryscinski-etal-2020-evaluating} (where the knowledge base is an input document or documents).
In this paper, we extend this definition to whether the claims made in \textbf{generated signals} (which could be text, code, or mathematical expressions and so on) can be supported by \textbf{evidence under specific rules}. Specifically, these rules can range from consistency with a knowledge base derived from Wikipedia, to a verification rule specified within a Python library, or an operational rule derived from mathematics. By adopting this broader definition, we are able to establish a unified framework for addressing factuality issues in generative AI beyond just the textual domain.

\paragraph{Fine-grained Factuality}
One can usually detect the factuality of a given generated signal (e.g., text) at different levels of granularity, such as sentences, and documents. A more granular assessment can be particularly valuable because it (1) not only allows users to pinpoint where inaccuracies occur~\cite{liu-etal-2021-explainaboard} but also (2) serves as a reward model for developers to refine their generative systems~\cite{lightman2023lets}.

However, implementing fine-grained factuality detection is challenging due to two reasons: (1) specifying the desired granularity level without ambiguity, and (2) extracting claims in line with the predetermined granularity level. In this paper, we argue that by utilizing the powerful \emph{instruction-following ability} and the \emph{natural language interface} of LLMs, we can effectively address the challenge of defining and extracting fine-grained claims through claim definition-based few-shot prompting. More details can be found in \S\ref{subsec:claim-extraction}.

Structurally speaking, given a prompt (e.g., a query or instruction) and the corresponding model-generated response, the fine-grained factuality detection task involves the following concepts:

\noindent \textbf{Prompt ($p$)} a query or instruction that users provide to the generative model.

\noindent \textbf{Response ($r$)} a piece of text (usually in long form) generated by the generative model.

\noindent \textbf{Claim ($c$)} a statement inferred from the model response, whose granularity is defined by a natural language text.

\noindent \textbf{Evidence ($e$)} The available information (e.g., knowledge base, pre-defined rules) that support or demonstrate the truth or validity of a claim.

\subsection{Instantiations in Different Scenarios}

Using the above task definition, we can define factuality in different application scenarios (see also in Tab.\ref{tab:factuality-definition}).

\paragraph{Knowledge-based QA}
Knowledge-based (KB) QA~\cite{chen-etal-2017-reading} aims to answer questions using a given knowledge base or open-domain data source (e.g., Wikipedia). In this task, we define factuality as how well each claim in the generated answer is supported by world knowledge. In this paper, we consider a more challenging scenario: open-domain QA that requires long-form answers, rather than short ones.

\paragraph{Code Generation}
The code generation task~\cite{yin-neubig-2017-syntactic} involves generating executable code based on a given query. We define factuality in code generation as how well the generated code, as a whole, can be executed correctly within a specific programming language (e.g., Python) and fulfills the provided requirements. This definition is grounded in an execution-based approach to code evaluation, which measures the correctness of generated code by executing it against some test case inputs and comparing its output to the expected output.

\paragraph{Math Problem Solving}

The math problem solving task involves the use of automated methods to address mathematical problems~\cite{cobbe2021training}. At the claim level, factuality in math problem solving is defined as the extent to which the generated statements adhere to the calculation rules. At the response level, factuality in math problem solving is defined as how effectively the overall mathematical solution addresses the given problem.

\paragraph{Scientific Literature Review Writing}
The scientific literature review writing task~\cite{jha2015surveyor} aims to analyze and synthesize existing research on a specific topic in a field of study.
In this task, we define factuality as whether the generated scientific literature review correctly cites existing scientific literature, including the correct mention of authors and publication years.\footnote{In this paper, our focus lies in examining the consistency of the relationship between the paper title, authors, and publication year. However, the task of determining the suitability of the cited paper as the most appropriate choice is left for future investigation.}

\begin{figure*}
    \centering
    \includegraphics[width=1\linewidth]{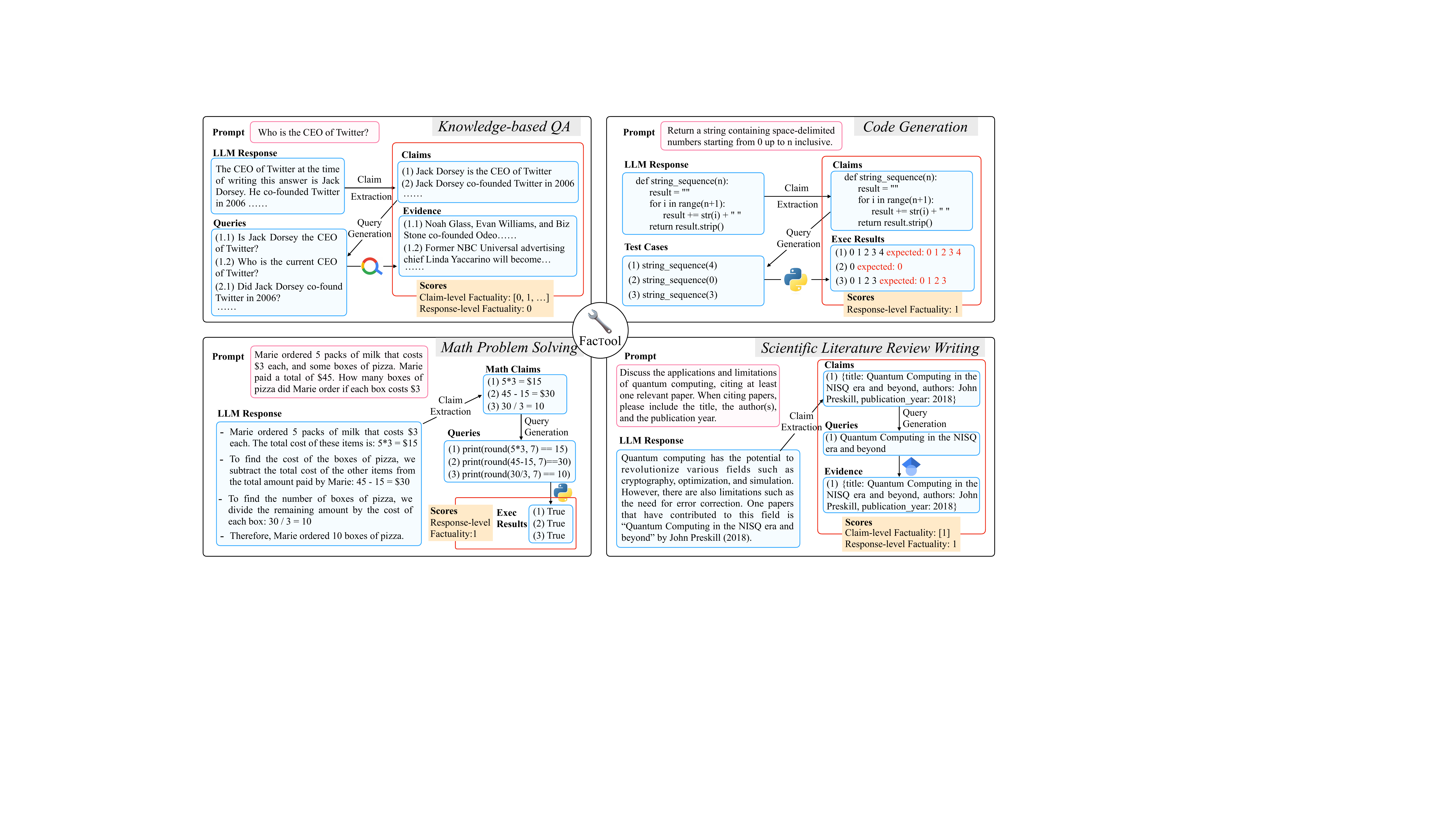}
    \caption{Our proposed framework for factuality detection in four domains: knowledge-based QA, code generation, math problem solving and scientific literature review writing. 
    }
    \label{fig:framework}
\end{figure*}

\section{Approach}

We propose a tool-augmented framework for detecting factual errors that can apply a unified approach across various tasks.
The motivation for using tools is twofold. On one hand, each tool embodies the domain expertise, assisting us in the effective gathering of evidence that verifies the correctness of the claim. On the other hand, the ability of LLMs to utilize multiple tools paves the way for \emph{multiple tool-augmented factuality detection}. For example, by directly using ChatGPT plugins,\footnote{\url{https://openai.com/blog/chatgpt-plugins}} we can integrate multiple tools into a chatbot.

The framework is illustrated in Fig.~\ref{fig:intro}, which consists of five main components:  \emph{claim extraction}, \emph{query generation}, \emph{tool querying}, \emph{evidence collection}, and \emph{agreement verification}. We elaborate each component below.

\subsection{Claim Extraction} \label{subsec:claim-extraction}
Extracting claims from responses under various task settings is challenging due to the inconsistent definitions of claims across tasks and scenarios. This inconsistency hinders the development of applications such as text summarization evaluation and factuality detection. 
To tackle this, we propose an approach in this paper that treats claim extraction as a process guided by LLM prompts based on the specific definition of claims. This approach offers the following advantages:

(i) Leveraging the strong instruction-following capabilities of LLMs can significantly reduce the costs associated with data annotation and model training for claim extraction.

(ii) When developing a system or constructing a dataset for an application that relies on the definition of claims, one simply needs to provide a textual definition of the claim using a large model. This enables future researchers to effectively utilize these definitions as a foundation in their work.

(iii) Our experiments demonstrate that the claim extraction module, implemented by ChatGPT, exhibits strong performance in extracting claims (atomic component units). The detailed results of these experiments are discussed in Section 6.1.

Here, we employ ChatGPT as a base LLM and apply different textual definitions of claims across four tasks. Our goal is to extract all verifiable claims within the generated text $x$, denoted as $\{c_i\}_{i = 1 \cdots n}$.
Detailed prompting instructions can be found in Appendix \ref{sec:appendix:a}.

\paragraph{KB-based QA}
The claim is defined using the concept of atomic content units (ACUs)~\cite{liu2022revisiting}. Each ACU corresponds to a single atomic fact within a generated answer.
In practice, we leverage ChatGPT\footnote{We have also explored other entailment-based models with BERT, and the result is no better than ChatGPT.} (specifically, the ``gpt-3.5-turbo'' version) to extract claims based on two criteria: (i) each claim should not exceed 15 words, and (ii) it should clearly describe a fact. 
We also include two in-context examples from the RoSE dataset~\cite{liu2022revisiting} in our prompt to obtain more fine-grained claims. Additionally, we ask ChatGPT to resolve any coreferences or ambiguity, such as unclear pronouns and other related expressions within the claims.

\paragraph{Code Generation}
We consider each generated code snippet within the response as a single claim to be verified. We extract all such code snippets that are enclosed with brackets, in other words, within a code block.

\paragraph{Math Problems}
We define each claim in a step-by-step math solution as the arithmetic operation performed between known real numbers. Each of these operations contains two parts: the calculation and the calculated answer. We prompt ChatGPT to extract all such claims.

\paragraph{Scientific Literature Review}
Each claim within the generated review is defined as a tuple of ``\textit{(paper title, year, authors)}'' contained from the generated review. We then prompt ChatGPT to extract all such tuples within the generated review.

\subsection{Query Generation}
For each claim $c_i$, we convert it into a list of queries $\{q_{ij}\}_{j = 1 \cdots m}$ that can be used to query external tools such as search engines, the Python interpreter, or Google scholar. 

\paragraph{KB-based QA}
We prompt ChatGPT or GPT-4 to generate two search engine queries from each claim $c_i$. These queries are intended to help humans in verifying the factuality of $c_i$. Detailed prompting instructions can be found in Appendix \ref{sec:appendix:a}.

\paragraph{Code Generation}
For each claim $c_i$ we generate two different types of queries: simulated test case inputs, denoted as $\{{q_t}_{ij}\}_{j = 1 \cdots m}$, and potential solutions, denoted as $\{{q_s}_{ij}\}_{j = 1 \cdots m}$. Both types of queries are generated by ChatGPT or GPT-4. The simulated test case inputs are function calls generated for a given code snippet, while potential solutions are repeatedly generated solutions that ChatGPT generates in response to the user prompt $p$. In our later experiments, we generate 3 simulated test case inputs and 3 potential solutions. Detailed prompting instructions can be found in Appendix \ref{sec:appendix:a}.

\paragraph{Math Problems}
We prompt ChatGPT or GPT-4 to convert all mathematical operations into executable Python code snippets. These snippets are designed to return ``True'' when the calculation matches the calculated answer and ``False'' if it doesn't. Detailed prompting instructions can be found in Appendix \ref{sec:appendix:a}.

\paragraph{Scientific Literature Review}
We use the paper title, found within the extracted claim tuple, as the query for Google Scholar. Our assumption here is that if a paper exists, it should appear as the first search result on Google Scholar when we use the paper title as the query.

\subsection{Tool Querying \& Evidence Collection}
\begin{figure}[t]
    \centering
    \includegraphics[width=1.0\linewidth]{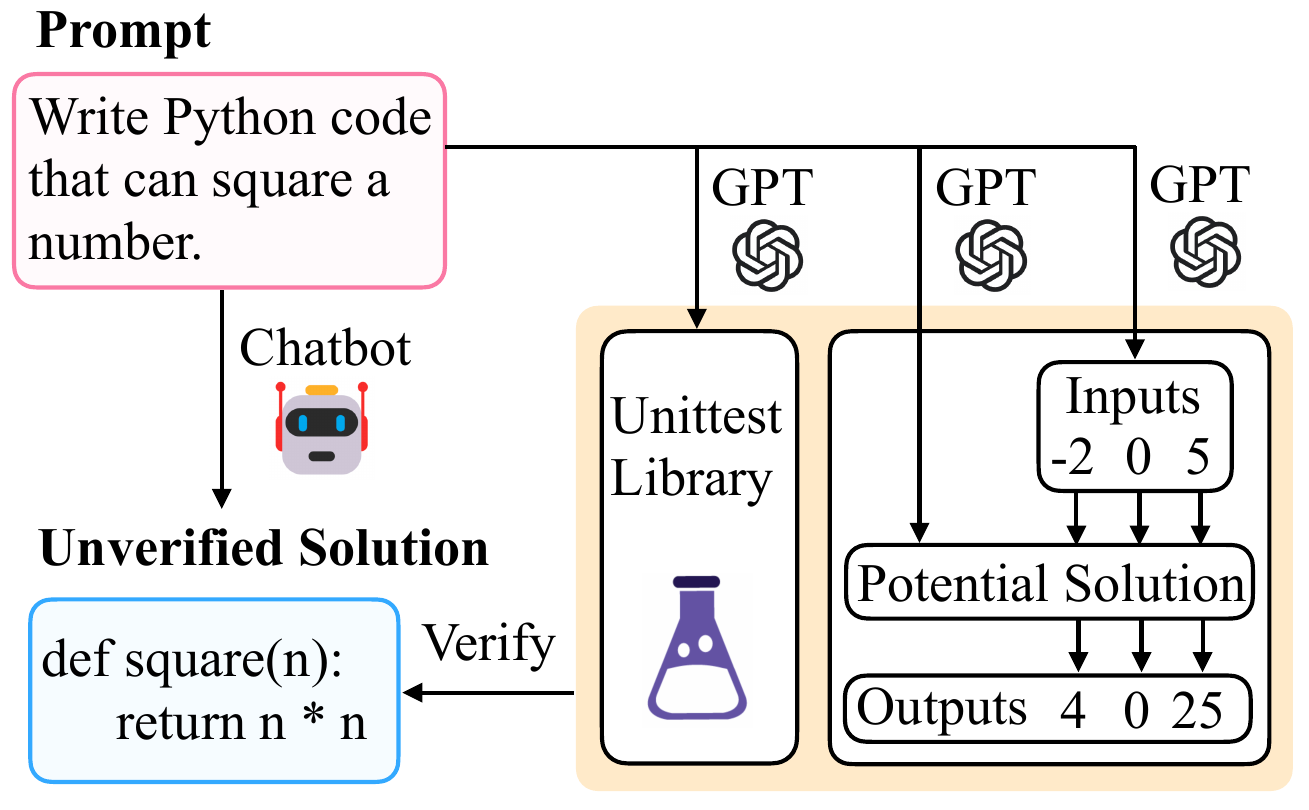}
    \caption{
    Unit test library generation for detecting factual errors in code generation.}
    \label{fig:unittest}
\end{figure}
We then use the queries to query various tools to collect relevant evidence statements $\{e_{ik}\}_{k = 1 \cdots l_i}$. 

\paragraph{KB-based QA}

The external tool we use to help verify the factuality of the generated text is the Google Search API, which queries the internet for knowledge using the queries generated from the claims extracted from the generated text of LLM.

We use the Google Search API provided by Serper\footnote{\url{https://serper.dev/}} to search the top pages and retrieve the most relevant search snippets included in the API's response. We then parse the response to obtain different types of snippets such as answer boxes, knowledge graphs, and organic search results.

\paragraph{Code Generation}
For each test case input $t_i$ and generated potential solution $s_j$, we execute $s_j$ using $t_i$ as the input and collect the execution result (output) for each $(t_i, s_j)$ pair. The input-output pairs are used as test cases for verifying the chatbot generated unverified solution. The process is shown in Fig.~\ref{fig:unittest}.

\paragraph{Math Problems}
We collect the execution results for code snippets derived from the mathematical operations. As illustrated in Fig.~\ref{fig:framework}, math claims like ``\texttt{30 /3 = 10}'' are extracted and then converted into a Python executable code, for instance, ``\texttt{print(round(30/3, 7)==10)}''.

\paragraph{Scientific Literature Review}
We use the title of each paper, extracted from the text, as the query to access relevant information through the Google Scholar API provided by the Scholarly\footnote{\url{https://github.com/scholarly-python-package/scholarly}} Python package. This allows us to retrieve key information about each paper, including the paper title, author list, and publication year.

\subsection{Agreement Verification}
In the final step, each claim, $c_i$, receives a binary factuality label, $L_{i} \in \{\textsc{True}, \textsc{False}\}$, based on the level of support it receives from the collected evidence, $\{e_{ik}\}_{k = 1 \cdots l_i}$. This labeling process is performed for every individual claim.

\paragraph{KB-based QA}
We prompt ChatGPT or GPT-4 to judge the factuality of the claim given the retrieved list of evidence snippets. We follow a zero-shot Chain-of-Thought~\cite{wei2023chainofthought} reasoning process: initially, the model attempts to reason about whether the claim is factual or not. If an error is identified, we then ask it to explain and attempt to rectify the mistake. 

\paragraph{Code Generation}
We conduct a majority vote for each test case across all solutions, establishing what we refer to as the ``pseudo-golden output'' for that particular test case. We repeat this process for every test case. Following this, we compare the execution result of the solution that’s under verification against all
the test cases with the pseudo golden output. If the results match, we classify the solution under
verification as true. Otherwise, it is deemed false.

\paragraph{Math Problems}
We compile the results of each code snippet execution. If any snippet returns ``False'', we classify the associated generated text $x$ as false. Conversely, if all snippets yield ``True'', we classify the corresponding generated text $x$ as true.

\paragraph{Scientific Literature Review}
We compare the extracted claim: ``\textit{(paper title, year, authors)}'' to the evidence: ``\textit{(paper title, year, authors)}'' retrieved from Google Scholar API. For the paper title and year of publication, we conduct an exact, case-insensitive string match. As for the authors' match, we prompt ChatGPT or GPT-4 to judge whether the author list in the extracted claim is a subset of the retrieved author list. All the information must be matched in order to be classified as ``True'', otherwise ``False''.

\section{Dataset Construction}

\subsection{Prompt and Response Collection}

\paragraph{KB-based QA}
For KB-based QA, we evaluate our framework using RoSE \cite{liu2022revisiting} and FactPrompts. RoSE is a text summarization dataset that provides fine-grained ACUs for each reference summary. FactPrompts is a dataset that comprises real-world prompts sourced from various platforms and datasets, such as Quora and TruthfulQA~\cite{lin-etal-2022-truthfulqa}, along with corresponding responses generated by ChatGPT. 
We construct the dataset using 100 reference summaries from RoSE and 50 responses from FactPrompts for our evaluation.

\paragraph{Code Generation}
For code generation, we evaluate our framework using HumanEval \cite{chen2021evaluating}. HumanEval is a programming problem dataset that contains several unit tests for each problem. We use ChatGPT to generate responses based on the processed prompts of HumanEval provided in \cite{chen2022codet} which solely contain the instruction of the prompt without input-output demonstrations.

\paragraph{Math Problems}
For math problems, we evaluate our framework using GSM-Hard \cite{gao2022pal}. GSM-Hard is a dataset constructed from GSM8K \cite{cobbe2021training} by replacing the numbers in the questions of GSM8K with larger numbers. We sampled 100 prompts from GSM-Hard that have a target solution value of positive.\footnote{GSM8K involves many application questions, including calculations involving money, measurements of quantities, etc. We found that GSM-Hard examples with negative values often contained illogical situations, such as ``negative 5 apples''. A positive target solution value helps prevent ChatGPT from making extra assumptions on top of the description in the problem.} 
Then, we generate responses for these prompts using ChatGPT.

\paragraph{Scientific Literature Review}
For the scientific literature review, we follow self-instruct~\cite{wang2023selfinstruct} to create 100 diverse prompts spanning computer science, business, law, medicine, and physics. Each prompt asks for a technical or research-oriented response that includes at least one relevant literature citation. Then, we generate responses for these prompts using ChatGPT.

\subsection{Claim Collection}
For responses from FactPrompts and GSM-Hard, we follow the idea of ``claim extraction as prompting'' described in \S\ref{subsec:claim-extraction}, This approach allows us to reuse claim prompts as listed in Appendix \ref{sec:appendix:a}.
We use ChatGPT as the model for claim extraction due to its cost efficiency and effectiveness in extracting fine-grained claims.
In terms of HumanEval responses, given that the generated response to a HumanEval prompt is already in the form of a code snippet, we consider the ``claim'' of the response to be identical to the response itself.

\subsection{Claim and Response Annotation}

\paragraph{KB-based QA \& Scientific Literature Review}
For claim annotation, the authors collectively annotate the extracted claims as either factual or non-factual. For response annotation, if one claim within the response is labeled as non-factual, then the response as a whole is considered non-factual; otherwise, the response is considered factual.

\paragraph{Code Generation}
We consider the claim label to be identical to the response label since the ``claim'' of the response is the same as the response itself. For response annotation, we annotate ChatGPT's responses using the execution code provided in \cite{chen2022codet} against the HumanEval test cases. This allows us to distinguish between factual (those passing all tests) responses and non-factual responses.

\paragraph{Math Problems}
For claim annotation, the authors collectively annotate the extracted claims as either factual or non-factual. For response annotation, we utilize the target value provided in GSM-Hard \cite{gao2022pal} to annotate the generated responses.

\begin{table}[htbp]
  \centering
  \footnotesize
  
    \begin{tabular}{cllrr}
    \toprule
    \textbf{Task} & \textbf{Datasets} & \multicolumn{1}{l}{\textbf{Responses}} & \multicolumn{1}{l}{\textbf{Claims}} \\
    \midrule
    KB-QA & RoSE  & \multicolumn{1}{l}{100}  &  \multicolumn{1}{l}{527}\\
    \midrule
    KB-QA & FactPrompts & \multicolumn{1}{l}{50 (23:27)} & 233 (177:56)\\
    Code  & HumanEval & \multicolumn{1}{l}{164 (109:55)} & 164 (109:55)\\
    Math  & GSM-Hard & \multicolumn{1}{l}{100 (47:53)}   &  284 (246:38)\\
    Sci.Lit & FactPrompts & \multicolumn{1}{l}{100 (10:90)}   &  186 (33:153)\\
    \bottomrule
    \end{tabular}%
    \caption{Detailed statistics of datasets used in this work. Note that (p, n) denotes p = count of positive responses or claims, and n = count of negative responses or claims. ``Sci.Lit'' represents scientific literature review.
    }
  \label{tab:data_stats}%
\end{table}%

\section{Experiments}
We evaluate \toolname against two baselines that use LLMs to check their own inputs: Self-Check with 3-shot CoT and zero-shot CoT, which have shown to been effective on various tasks including dialogue response, math reasoning, and code generation \cite{madaan2023selfrefine,chen2023teaching}. Both of these baselines aim to test the ability of LLM to identify its own errors without the use of any external tool.
In practice, we prompt ChatGPT (gpt-3.5-turbo-0301) and GPT-4 (gpt-4-0314)\footnote{We anticipate that the recently released models, gpt-3.5-turbo-0613 and gpt-4-0613, will lower the inference costs for \toolname. This expectation arises from their improved ability to produce structured responses, such as those in JSON format. While conducting our experiments on gpt-3.5-turbo-0301 and gpt-4-0314, we often ran into problems where the responses were not valid JSON, requiring us to rerun any samples with invalid response formats. The source code of \toolname will be using the latest versions of ChatGPT and GPT-4.}
to recognize, explain, and attempt to rectify their own errors. Following this reasoning process, the models make final judgments on the factuality of the given claim. The key difference between Self-Check (zero-shot CoT) and Self-Check (3-shot CoT) is that Self-Check (3-shot CoT) provides three demonstrations to models, while Self-Check (zero-shot CoT) does not provide any demonstrations.

\subsection{Exp-I: Claim Extraction Evaluation}
We evaluate the claim extraction module of \toolname on RoSE  \cite{liu2022revisiting}. We treat the reference summary as the generated text $x$, and the reference ACUs as the golden-extracted claims. We measure the similarity between the machine-extracted (GPT-4, ChatGPT, and Flan-T5 XXL) claims $\{c^c_i\}_{i = 1 \cdots n_c}$ and golden-extracted claims $\{c^g_i\}_{i = 1 \cdots n_g}$ using 4 metrics: ROUGE-1, ROUGE-2, ROUGE-L~\cite{lin-2004-rouge}, and BERTScore. In Tab.~\ref{tab:rose_result}, we report the average of the highest similarity between each ChatGPT-extracted claim and the corresponding golden-extracted claim in the same sample. (i.e., $\frac{1}{\text{sample\_cnt}}\sum_{\text{sample}}\frac{1}{n_c}\sum_{i=1}^{n_c} \max_{j=1}^{n_g} (\mathrm{{Sim}}(c^c_i, c^g_j))$).

\begin{table}[!t]
\centering
\scriptsize
\begin{tabular}{@{}llccc@{}}
\toprule
Model & Metric & Precision & Recall & F1-score \\ \midrule
GPT-4 & ROUGE-1 & 0.7394 & 0.8758 & \textbf{0.7860} \\
 & ROUGE-2 & 0.6304 & 0.7771 & \textbf{0.6772} \\
 & ROUGE-L & 0.7175 & 0.8625 & \textbf{0.7667} \\
 & BERTScore & 0.6632 & \textbf{0.7865} & \textbf{0.7175} \\ 
\midrule
 ChatGPT & ROUGE-1 & \textbf{0.7770} & 0.8285 & 0.7836 \\
 & ROUGE-2 & \textbf{0.6520} & 0.7115 & 0.6610 \\
 & ROUGE-L & \textbf{0.7557} & 0.8148 & 0.7655 \\
 & BERTScore & \textbf{0.6958} & 0.7521 & 0.7174 \\ 
\midrule
 FLAN-T5-XXL & ROUGE-1 & 0.6531 & \textbf{0.8928} & 0.7326 \\
 & ROUGE-2 & 0.5609 & \textbf{0.8157} & 0.6413 \\
 & ROUGE-L & 0.6428 & \textbf{0.8885} & 0.7237 \\
 & BERTScore & 0.4314 & 0.6661 & 0.5408 \\ 
\bottomrule
\end{tabular}
\caption{The average similarity between the extracted claims from GPT-4, ChatGPT, and Flan-T5 XXL and the golden ACUs on RoSE. 
}
\label{tab:rose_result}
\end{table}

\paragraph{Results}
We demonstrate in Tab.~\ref{tab:rose_result}
that the claims extracted by GPT-4, ChatGPT, and Flan-T5 closely match the ACUs annotated by humans, as evaluated by ROUGE and BERTScore metrics. Note that in Exp-II, we choose ChatGPT as the claim extractor for two reasons: (1) The context length of Flan-T5 is too short (512 tokens) to effectively extract claims from lengthy responses in our dataset. (2) ChatGPT is more cost-efficient compared to GPT-4, while maintaining similar effectiveness in claim extraction.

\subsection{Exp-II: Framework Evaluation}
We evaluate \toolname and the two Self-Check baselines on the dataset constructed from each scenario. Depending on the model used for query generation and agreement verification, we have two \toolname baselines: \toolname powered by ChatGPT and \toolname powered by GPT-4. We report the accuracy, recall, precision, and F1-score at both the claim and response levels.

\subsubsection{Result}

\begin{table*}[!htbp]
  \centering
  \footnotesize
    \begin{tabular}{crccccc|cccc}
    \toprule
    \multicolumn{1}{r}{\multirow{2}[4]{*}{\textbf{Tasks}}} & \multicolumn{1}{r}{\multirow{2}[4]{*}{\textbf{LLMs}}} & \multirow{2}[4]{*}{\textbf{Methods}} & \multicolumn{4}{c}{\textbf{Claim-Level}} & \multicolumn{4}{c}{\textbf{Response-Level}} \\
\cmidrule{4-11}          &       &       & \textbf{Acc.} & \textbf{R} & \textbf{P} & \textbf{F1} & \textbf{Acc.} & \textbf{R} & \textbf{P} & \textbf{F1} \\
    \midrule
    \multicolumn{1}{r}{\multirow{6}[4]{*}{KB-QA}} & \multicolumn{1}{r}{\multirow{3}[2]{*}{ChatGPT}} & Self-Check (0) & 75.54 & \textbf{90.40} & 80.00 & 84.88 & 54.00 & 60.87 & 50.00 & 54.90 \\
          &       & Self-Check (3) & 69.53 & 81.36 & 79.12 & 80.23 & 54.00 & 47.83 & 50.00 & 48.89 \\
          &       & \textbf{\toolname} & 74.25 & 73.45 & 90.91 & 81.25 & 64.00 & 43.48 & 66.67 & 52.63 \\
\cmidrule{2-11}          & \multicolumn{1}{r}{\multirow{3}[2]{*}{GPT-4}} & Self-Check (0) & 77.25 & 84.75 & 85.23 & 84.99 & 54.00 & \textbf{95.65} & 50.00 & 65.67 \\
          &       & Self-Check (3) & 79.83 & 85.88 & 87.36 & 86.61 & 64.00 & 52.17 & 63.16 & 57.14 \\
          &       & \textbf{\toolname} & \textbf{84.12} & 85.31 & \textbf{93.21} & \textbf{89.09} & \textbf{78.00} & 60.87 & \textbf{87.50} & \textbf{71.79} \\
    \midrule
    \multicolumn{1}{r}{\multirow{6}[4]{*}{Code}} & \multicolumn{1}{r}{\multirow{3}[2]{*}{ChatGPT}} & Self-Check (0) & 68.29 & 99.10 & 68.33 & 80.88 & 68.29 & 99.10 & 68.33 & 80.88 \\
          &       & Self-Check (3) & 68.90 & \textbf{100.00} & 68.52 & 81.32 & 68.90 & \textbf{100.00} & 68.52 & 81.32 \\
          &       & \toolname & 78.05 & 89.19 & 80.49 & 84.62 & 78.05 & 89.19 & 80.49 & 84.62 \\
\cmidrule{2-11}          & \multicolumn{1}{r}{\multirow{3}[2]{*}{GPT-4}} & Self-Check (0) & 75.31 & 95.50 & 75.18 & 84.13 & 75.31 & 95.50 & 75.18 & 84.13 \\
          &       & Self-Check (3) & 77.44 & 96.40 & 76.43 & 85.26 & 77.44 & 96.40 & 76.43 & 85.26 \\
          &       & \textbf{\toolname} & \textbf{89.02} & 94.59 & \textbf{89.74} & \textbf{92.11} & \textbf{89.02} & 94.59 & \textbf{89.74} & \textbf{92.11} \\
    \midrule
    \multicolumn{1}{r}{\multirow{6}[4]{*}{Math}} & \multicolumn{1}{r}{\multirow{3}[2]{*}{ChatGPT}} & Self-Check (0) & 84.15 & 90.24 & 91.36 & 90.80 & 57.00 & 74.47 & 53.03 & 61.95 \\
          &       & Self-Check (3) & 87.32 & 94.31 & 91.34 & 92.80 & 61.00 & 89.36 & 55.26 & 68.29 \\
          &       & \toolname & 97.54 & 97.56 & 99.59 & 98.56 & \textbf{78.00} & 93.62 & \textbf{69.84} & 80.00 \\
\cmidrule{2-11}          & \multicolumn{1}{r}{\multirow{3}[2]{*}{GPT-4}} & Self-Check (0) & 83.10 & 86.99 & 93.04 & 89.92 & 49.00 & 85.11 & 47.62 & 61.07 \\
          &       & Self-Check (3) & 92.61 & 96.75 & 94.82 & 95.77 & 65.00 & 89.36 & 58.33 & 70.59 \\
          &       & \textbf{\toolname} & \textbf{98.24} & \textbf{97.97} & \textbf{100.00} & \textbf{98.97} & \textbf{78.00} & \textbf{95.74} & 69.23 & \textbf{80.36} \\
    \midrule
    \multicolumn{1}{r}{\multirow{6}[4]{*}{Scientific}} & \multicolumn{1}{r}{\multirow{3}[2]{*}{ChatGPT}} & Self-Check (0) & 28.69 & 96.00 & 21.82 & 35.56 & 18.00 & \textbf{100.00} & 10.87 & 19.61 \\
          &       & Self-Check (3) & 24.19 & \textbf{96.97} & 18.60 & 31.22 & 22.00 & 90.00 & 10.47 & 18.75 \\
          &       & \toolname & 97.31 & 84.85 & \textbf{100.00} & 91.80 & \textbf{99.00} & 90.00 & \textbf{100.00} & \textbf{94.74} \\
\cmidrule{2-11}          & \multicolumn{1}{r}{\multirow{3}[2]{*}{GPT-4}} & Self-Check (0) & 35.75 & 84.85 & 20.29 & 32.75 & 19.00 & \textbf{100.00} & 10.99 & 19.80 \\
          &       & Self-Check (3) & 44.75 & 87.88 & 23.20 & 36.71 & 49.00 & 70.00 & 12.73 & 21.54 \\
          &       & \textbf{\toolname} & \textbf{98.39} & 90.91 & \textbf{100.00} & \textbf{95.24} & \textbf{99.00} & 90.00 & \textbf{100.00} & \textbf{94.74} \\
    \bottomrule
    \end{tabular}%
      \caption{Experimental results of \toolname powered by ChatGPT and \toolname powered by GPT-4 on KB-based QA, Code Generation, Math Problems, and Scientific Literature Review.
      }
  \label{tab:allresults}%
\end{table*}%

Tab.~\ref{tab:allresults} shows the claim-level and response-level performance of \toolname and the self-check baselines. We obtain following observations.

\paragraph{\toolname powered by GPT-4 outperforms all other baselines across all scenarios}
From Tab.~\ref{tab:allresults}, we observe that \toolname powered by GPT-4 outperforms all other baselines across all scenarios. \toolname powered by GPT-4 achieves an $89.09$ claim-level F1 / $71.79$ response-level F1 on KB-based QA, a $92.11$ claim-level F1 / $92.11$ response-level F1 on code generation (remember that claim-level factuality is considered equivalent to response-level factuality in our experiment for code generation), a $98.97$ claim-level F1 / $80.36$ response-level F1 on math problems, and a $95.24$ claim-level F1 / $94.74$ response-level F1 on scientific literature review. Each of these figures is the highest for their respective tasks.

\paragraph{\toolname powered by GPT-4 outperforms all self-check baselines across all scenarios}
From Tab.~\ref{tab:allresults}, we show that \toolname with GPT-4 outperforms all self-check baselines across all scenarios. On \toolname powered by GPT-4 v.s. Self-Check (3) powered by GPT-4, we observe: $71.79$ v.s. $57.14$ response-level F1 on KB-based QA, $92.11$ v.s. $85.26$ response-level F1 on code generation, $80.36$ v.s. $70.59$ response-level F1 on math problems, and $94.74$ v.s. $21.54$ response-level F1 on scientific literature review.

\paragraph{\toolname powered by GPT-4 significantly outperforms all self-check baselines in scientific literature review}
From Tab.~\ref{tab:allresults}, we show that \toolname powered by GPT-4 significantly outperforms the self-check baselines in scientific literature review. On \toolname powered by GPT-4 v.s. Self-Check (3) powered by GPT-4, we observe: $95.24$ v.s. $36.71$ claim-level F1 and $94.74$ v.s. $21.54$ response-level F1. Here, Google Scholar shown to be highly robust in performing its specified task of finding citations when compared to LLM itself.

\paragraph{\toolname powered by GPT-4 outperforms \toolname powered by ChatGPT}
\toolname powered by GPT-4 outperforms \toolname powered by ChatGPT across all scenarios. This trend is especially significant in KB-QA, where query generation and agreement verification are harder for ChatGPT but relatively easier for GPT-4 ($89.09$ v.s $81.25$ claim-level F1 and $71.79$ v.s $52.63$ response-level F1). On the other hand, in scenarios where query generation and agreement verification are relatively easy for both ChatGPT and GPT-4, the performance is similarly good.

\paragraph{Self-check models are prone to false positive and thus less sensitive in detecting errors}
From Tab.~\ref{tab:allresults}, we observe that self-check models have lower precision compared to \toolname. On Self-Check (3) powered by GPT-4 v.s. \toolname powered by GPT-4, we observe: $63.16$ v.s. $87.50$ response-level precision on KB-based QA, $76.43$ v.s. $89.74$ response-level precision on code generation, $58.33$ v.s. $69.23$ response-level precision on math problems, and $12.73$ v.s. $100.00$ response-level precision on scientific literature review. These figures show that self-check models tend to classify claims as ``True'' considerably more frequently than \toolname, suggesting a lower sensitivity for error detection. 

\paragraph{Self-check models powered by ChatGPT outperform \toolname powered by ChatGPT on KB-QA}
Tab.~\ref{tab:allresults} shows that Self-Check (0) powered by ChatGPT outperforms \toolname powered by ChatGPT. Through examining specific cases, we found that reasoning errors are the main reason why \toolname powered by ChatGPT performs worse than the self-check baselines. Even when provided with sufficient evidence to determine whether the claim is factual or not, the agreement verification implemented by ChatGPT can become confused. For example, for the claim ``\texttt{The modern-day version of fortune cookies was invented in the United States.}'', the reasoning of \toolname powered by ChatGPT is self-contradictory: ``\texttt{The given text is not entirely factual. The modern-day version of fortune cookies was not invented in the United States. Most people nowadays believe that fortune cookies were created by a Japanese man named Makoto Hagiwara in 1914 in San Francisco. Hagiwara owned what is now called the Golden Gate Park Japanese Tea Garden, where he served tea and fortune cookies. This is supported by the provided evidences.}''
Detailed examples can be found in Fig.~\ref{fig:eval_prompt_single} of Appendix \ref{sec:appendix:b}.

\subsubsection{Performance Analysis}
We take a closer look at performance in different scenarios by examining evaluated cases.

\paragraph{KB-based QA}

The fact-checking capability of \toolname on KB-based QA is determined by several factors, including whether the search engine can return the most relevant snippets that could assist in determining the factuality of the given claim, the quality of the generated search engine queries, and the LLM's ability to reason about the validity of the claim given the retrieved evidence.
We found that \toolname powered by GPT-4 is especially capable under the following situations:
(1) Fact-checking recent events, discoveries, or news: \toolname powered by GPT-4 successfully identify false claims such as ``\texttt{Argentina has not won the World Cup since 1986}'' and ``\texttt{The most valuable NFT ever sold is a digital artwork called `Everydays: The First 5000 Days'}''.
(2) Fact-checking high-precision statistics: \toolname powered by GPT-4 successfully identify false claims such as ``\texttt{Ireland has an obesity rate of 26.9\%}'' and ``\texttt{Everydays: The First 5000 Days' sold for $69$ million}''. Detailed examples can be found in Fig.~\ref{fig:KB-QA_example_1} of Appendix \ref{sec:appendix:b}.

\paragraph{Code Generation}
The fact-checking capability of \toolname on code generation is determined by the LLM's capability to generate high-quality test cases and potential solutions. We demonstrate that due to GPT-4's exceptional ability to generate such high-quality test cases and potential solutions, \toolname powered by GPT-4 outperforms other baselines. For example, in ``\texttt{HumanEval/36}'', GPT-4 is consistently generating high quality solutions, leading to its correctly identifies the mistakes in the response, while ChatGPT fails to identify the mistake. Detailed examples can be found in Fig.~\ref{fig:code_example_1} and Fig.~\ref{fig:code_example_2} of Appendix \ref{sec:appendix:b}.

\paragraph{Math Problems}
The fact-checking capability of \toolname on math problems is determined by the LLM's capability to generate accurate Python snippets that verify the correctness of given extracted mathematical calculations. Both \toolname powered by GPT-4 and \toolname powered by ChatGPT excel in this regard. For example, both \toolname powered by GPT-4 and \toolname powered by ChatGPT correctly identify $23 \times 4319216$ doesn't equal to $99305768$. Detailed examples can be found in Fig.~\ref{fig:math_example} of Appendix \ref{sec:appendix:b}. 

\paragraph{Scientific Literature Review}
The fact-checking capability of \toolname on Scientific Literature Review is determined by the LLM's capability to identifying whether the author list generated is a subset of the actual author list. Both \toolname powered by GPT-4 and \toolname powered by ChatGPT excel in this regard. 
For example, both \toolname powered by GPT-4 and \toolname powered by ChatGPT correctly identify that the paper ``\texttt{The Impact of Artificial Intelligence on Employment}'' was not written by ``\texttt{Acemoglu and Restrepo}''. Detailed examples can be found in Fig.~\ref{fig:scientific_example} of Appendix \ref{sec:appendix:b}.

\subsubsection{Failure Analysis}
To gain a comprehensive understanding of \toolname's performance, we conduct analysis on cases where \toolname will fail. 

\paragraph{KB-based QA}
We summarize following sources of errors:
(1) Reasoning error: Although the evidence provided is sufficient and the LLM accurately finds the most relevant information, the model fails to reason about the relationship between the claim and the provided evidence. For example, for claim ``\texttt{Jupiter is less dense than Saturn}", \toolname powered by GPT-4 fails to reason the relative relationship even though the evidences provided are sufficient. (2) Conflicting evidence: Conflict in evidence can cause confusion for LLM, leading to incorrect decisions. For example, for claim ``\texttt{Jupiter has a density of 1.33 grams per cubic centimeter}", there are conflicting evidences claiming that the density is 1.326 or 1.33g/cm$^3$ .
(3) Ambiguity in claim: Ambiguous descriptions and subjective adjectives can lead to incorrect decisions. For example, the claim ``\texttt{Fortune cookies are enjoyed by people all over the world.}" is ambiguous and can have different answers based on different interpretations. Detailed examples can be found in Fig.~\ref{fig:kbqa_error} of Appendix \ref{sec:appendix:b}.

\paragraph{Code Generation}
Errors in code generation mainly comes from:
(1) Limited variety in synthetic test cases: The synthetic test cases generated by LLMs may not be fully representative or sufficiently diverse. For example, in the ``\texttt{HumanEval/64}'' sample, all the inputs of the generated synthetic test cases are composed of strings that only include lowercase letters (without uppercase letters). 
(2) Potential errors in code generation: The generated potential solutions could contain errors or bugs. Despite implementing a majority voting system to lessen this issue, it cannot completely eliminate the chance of bugs in the code generation process. For example, in the ``\texttt{HumanEval/79}'' sample, all the generated solutions failed to correctly ``\texttt{decimal\_to\_binary(0)}'' as ``\texttt{db0db}''. Detailed examples can be found in Fig.~\ref{fig:code_error} of Appendix \ref{sec:appendix:b}.

\paragraph{Math Problems}
There are two major types of errors in factuality detection for math problems:
(1) Round-off error: Round-off errors can occur during numerical calculations in Python. For example, \toolname powered by GPT-4 incorrectly classify the math calculation ``\texttt{60444034 / 12 = 5037002.83}'' as ``\texttt{False}''. (2) Reasoning error: Since the claims extracted by \toolname only involve mathematical calculations, \toolname will not verify the reasoning process of the mathematical solution. For example, for the question ``\texttt{Kylar went to the store to buy glasses for his new apartment. One glass costs \$5, but every second glass costs only 60\% of the price. Kylar wants to buy 5364765 glasses. How much does he need to pay for them?}'', the ChatGPT generated response contains reasoning error that incorrectly substitute the total cost as ``\texttt{5,364,765 * 5}''. However, since \toolname only checks math calculation errors, \toolname powered by GPT-4 did not identify the reasoning error. Detailed examples can be found in Fig.~\ref{fig:math_error} of Appendix \ref{sec:appendix:b}.

\paragraph{Scientific Literature Review}
There are two major types of errors in factuality detection for scientific literature review:
(1) Errors in title matching: Title matching can sometimes be problematic due to abbreviations in the generated citations or the retrieved title. For example, although the paper ``\texttt{MDMA-assisted psychotherapy for treatment of PTSD: study design and rationale for phase 3 trials based on pooled analysis of six phase 2 randomized controlled trials} exists, \toolname powered by GPT-4 identify the paper title as incorrect. (2) Errors in author matching: the author matching process might sometimes not be robust. For example, although the authors of ``\texttt{Language Models are Unsupervised Multitask Learners"} are indeed
``\texttt{Alec Radford, Jeffrey Wu, Rewon Child, David Luan, Dario Amodei, and Ilya Sutskever}, \toolname powered by GPT-4 identify the author list as incorrect. Detailed examples can be found in Fig.~\ref{fig:scientific_error} of Appendix \ref{sec:appendix:b}.

\begin{figure*}[!ht]
    \centering
    \begin{minipage}{0.48\textwidth}
        \centering
        \footnotesize
        \resizebox{\textwidth}{!}{




\definecolor{palemagenta}{rgb}{0.98, 0.52, 0.9}


\begin{tikzpicture}
  \centering
  \begin{axis}[ybar=1.0pt,
    height=6cm,
    width=15cm,
    bar width=0.25cm,
    ymin=-0.3,
    enlarge y limits={upper,value=0.15},
    axis lines*=left,
    legend style={font=\normalsize,at={(0.5,1.1)},anchor=north,legend columns=-1,
    /tikz/every even column/.append style={column sep=0.1cm}},
    ylabel={Accuracy},
    label style={font=\LARGE},
    xticklabels={KB-QA, Code, Math, Scientific, TAC-2009},
    xtick={0,...,5},
    xtick=data,
    xmajorgrids=true,
    ymajorgrids=true,
    zmajorgrids=true,
    grid style=dashed,
    xticklabel style={
        font=\large,
        inner sep=0.2pt,
    },
    ]
    \addplot [draw=black!100,fill=cyan!60] table[x index=0,y index=1]{./figures/claim_acc_distribution.txt};
    \addplot [draw=black!100,fill=blue!60] table[x index=0,y index=2]{./figures/claim_acc_distribution.txt};
    \addplot [draw=black!100,fill=yellow!60] table[x index=0,y index=3] {./figures/claim_acc_distribution.txt};
    \addplot [draw=black!100,fill=red!60] table[x index=0,y index=4] {./figures/claim_acc_distribution.txt};
    \addplot [draw=black!100,fill=orange!60] table[x index=0,y index=5] {./figures/claim_acc_distribution.txt};    
    \legend{GPT-4, ChatGPT, Bard, Claude-v1, Vicuna-13B}
  \end{axis}
\end{tikzpicture}

}
        \caption{Claim-Level Accuracy across scenarios for GPT-4, ChatGPT, Bard, Claude-v1, and Vicuna-13B}
        \label{fig:claim-level}
    \end{minipage}
    \hfill
    \begin{minipage}{0.48\textwidth}
        \centering
        \footnotesize
        \resizebox{\textwidth}{!}{




\definecolor{palemagenta}{rgb}{0.98, 0.52, 0.9}


\begin{tikzpicture}
  \centering
  \begin{axis}[ybar=1.0pt,
    height=6cm,
    width=15cm,
    bar width=0.25cm,
    ymin=-0.003,
    enlarge y limits={upper,value=0.15},
    axis lines*=left,
    legend style={font=\normalsize,at={(0.5,1.1)},anchor=north,legend columns=-1,
    /tikz/every even column/.append style={column sep=0.1cm}},
    ylabel={Accuracy},
    label style={font=\LARGE},
    xticklabels={KB-QA, Code, Math, Scientific, TAC-2009},
    xtick={0,...,5},
    xtick=data,
    xmajorgrids=true,
    ymajorgrids=true,
    zmajorgrids=true,
    grid style=dashed,
    xticklabel style={
        font=\large,
        inner sep=0.2pt,
    },
    ]
    \addplot [draw=black!100,fill=cyan!60] table[x index=0,y index=1]{./figures/res_acc_distribution.txt};
    \addplot [draw=black!100,fill=blue!60] table[x index=0,y index=2]{./figures/res_acc_distribution.txt};
    \addplot [draw=black!100,fill=yellow!60] table[x index=0,y index=3] {./figures/res_acc_distribution.txt};
    \addplot [draw=black!100,fill=red!60] table[x index=0,y index=4] {./figures/res_acc_distribution.txt};
    \addplot [draw=black!100,fill=orange!60] table[x index=0,y index=5] {./figures/res_acc_distribution.txt};    
    \legend{GPT-4, ChatGPT, Bard, Claude-v1, Vicuna-13B}
  \end{axis}
\end{tikzpicture}

}
        \caption{Response-Level Accuracy across scenarios for GPT-4, ChatGPT, Bard, Claude-v1, and Vicuna-13B}
        \label{fig:response-level}
    \end{minipage}
\end{figure*}

\subsection{Exp-III: Using \toolname to Evaluate the Factuality of Modern Chatbots}
The purpose of developing a factuality detector is to audit the actual generative chatbots to assess the reliability of the responses generated by chatbots. To this end, we evaluate the factuality of modern chatbots, including GPT-4, ChatGPT, Claude-v1, Bard, and Vicuna-13B, using \toolname powered by GPT-4. It is important to note that in Exp-III, we consider \toolname as a golden evaluator, responsible for evaluating the factual accuracy of the responses generated by different chatbots. For prompts selection, we follow the same intuition as \cite{zhou2023lima}: KB-QA is the most common scenario. Thus, we select 30 KB-QA prompts, 10 code prompts, 10 math prompts. and 10 scientific prompts (i.e., 3 times more KB-QA prompts compare to prompts from other scenarios) to carry out this factuality evaluation on chatbots. The KB-QA prompts are collected from \cite{zhou2023lima}, code prompts from HumanEval \cite{chen2021evaluating}, math prompts from \cite{gao2022pal}, while the scientific prompts are generated by us. Responses for these prompts are generated by each of the evaluated chatbots.

We report both the claim-level and response-level accuracies for each chatbot, as evaluated by \toolname powered by GPT-4. Given that KB-QA responses contain significantly more claims than responses from other scenarios, we report the weighted claim-level accuracy. This weight is determined by the ratio of the number of prompts in each scenario. In other words, \\
\begin{equation*}
\resizebox{0.4\textwidth}{!}{%
$\begin{aligned}
&\mathbf{weighted\_claim\_level\_accuracy} \\
&= \frac{3}{6} \times \mathit{claim\_level\_accuracy \, in \, KB-QA} \\
&\phantom{=} + \frac{1}{6} \times \mathit{claim\_level\_accuracy \, in \, Code} \\
&\phantom{=} + \frac{1}{6} \times \mathit{claim\_level\_accuracy \, in \, Math} \\
&\phantom{=} + \frac{1}{6} \times \mathit{claim\_level\_accuracy \, in \, Scientific}
\end{aligned}$
}
\end{equation*}

Adopting the weighted-claim level accuracy evaluation helps us provide a more holistic and fair assessment of each chatbot's factual accuracy.

\paragraph{Results}

Tab.~\ref{tab:chatbot_eval} shows that GPT-4 has the best weighted claim-level factual accuracy and response-level accuracy compared to ChatGPT, Bard, Claude-v1, and Vicuna. 
Fig.~\ref{fig:claim-level} and \ref{fig:response-level} demonstrate fine-grained performance w.r.t each scenario (KB-QA, code, math, scientific).
We observe that (a) GPT-4 has the best claim-level accuracy and response-level accuracy in most of the scenarios. (b) Supervised fine-tuned Chatbots like Vicuna-13B perform reasonably well in more common scenarios like KB-QA but less so in more challenging scenarios such as math, code, and scientific.

\begin{table}[htbp]
\centering
\footnotesize
\begin{tabular}{@{}lccc@{}}
\toprule
LLMs & WCL Acc. & RL Acc. & Avg. Resp. Len. \\ \midrule
GPT-4 & \textbf{75.60} & \textbf{43.33} & 196.83 \\
ChatGPT & 68.63 & 36.67 & 144.05 \\
Claude-v1 & 63.95 & 26.67 & 208.70 \\ 
Bard & 61.15 & 33.33 & \textbf{263.77} \\
Vicuna-13B & 50.35 & 21.67 & 207.13 \\
\bottomrule
\end{tabular}
\caption{Factual accuracy of different chatbots evaluated by \toolname. WCL Acc. stands for weighted claim-level accuracy of each chatbot. RL Acc. stands for response-level accuracy of each chatbot. Avg. Resp. Len. stands for average response length of each chatbot. Note that we consider \toolname as the golden evaluator that evaluates the factuality of the responses generated by each chatbot.
}
\label{tab:chatbot_eval}
\end{table}

\section{Conclusion}
We introduce \toolname, a task- and domain-agnostic framework designed to tackle the escalating challenge of factual error detection in generative AI. We expand the conventional definition of factuality, particularly focusing on auditing the capabilities of generative AI models.  Realizing that (1) the generated texts of LLMs tend to be lengthy and lack a clearly defined granularity for individual facts, and that (2) there is a scarcity of explicit evidence available during the process of fact checking, we build \toolname as a 5-step tool-augmented framework that consists of claim extraction, query generation, tool querying, evidence collection, and verification. 

We demonstrate the potential of incorporating tools like Google Search, Google Scholar, code interpreters, Python, and even LLMs themselves in factual error detection through experimentation across diverse tasks such as knowledge-based QA, code generation, math problem solving, and scientific literature review writing. We believe that our holistic and adaptable framework can be easily extended to more scenarios.

\section*{Acknowledgements}
We thank Yixin Liu, Zhengbao Jiang, Zhiruo Wang for the useful discussion and suggestions.

\bibliography{anthology,custom}
\bibliographystyle{acl_natbib}

\cleardoublepage

\appendix

\section{Prompts}

We list the claim extraction, query generation, and agreement verification prompts used in this paper. All the prompts listed are user prompts. We use the same system prompt ``You are a brilliant assistant.''

\label{sec:appendix:a}

\begin{figure*}[htbp]
    \scriptsize
    \centering
\begin{tabular}
{@{}p{0.33\textwidth}p{0.33\textwidth}p{0.33\textwidth}@{}}
\toprule 
\textbf{[KB-Based QA]}

You are given a piece of text that includes knowledge claims. A claim is a statement that asserts something as true or false, which can be verified by humans. 

\textbf{[Task]}

Your task is to accurately identify and extract every claim stated in the provided text. Then, resolve any coreference (pronouns or other referring expressions) in the claim for clarity. Each claim should be concise (less than 15 words) and self-contained.


Your response MUST be a list of dictionaries. Each dictionary should contains the key "claim", which correspond to the extracted claim (with all coreferences resolved). You MUST only respond in the format as described below. DO NOT RESPOND WITH ANYTHING ELSE. ADDING ANY OTHER EXTRA NOTES THAT VIOLATE THE RESPONSE FORMAT IS BANNED. START YOUR RESPONSE WITH '['.

    \textbf{[Response Format]}

    [\{"claim": "Ensure that the claim is fewer than 15 words and conveys a complete idea. Resolve any coreference (pronouns or other referring expressions) in the claim for clarity."
      \},...
    ]

    \textbf{Here are two examples}:
    
    [\textbf{text}]: 
    
    Tomas Berdych defeated Gael Monfis 6-1, 6-4 on Saturday. The sixth-seed reaches Monte Carlo Masters final for the first time . Berdych will face either Rafael Nadal or Novak Djokovic in the final.
    
    [\textbf{response}]: 
    
    [\{"claim": "Tomas Berdych defeated Gael Monfis 6-1, 6-4"\}, \{"claim": "Tomas Berdych defeated Gael Monfis 6-1, 6-4 on Saturday"\}, \{"claim": "Tomas Berdych reaches Monte Carlo Masters final"\}, \{"claim": "Tomas Berdych is the sixth-seed"\}, \{"claim": "Tomas Berdych reaches Monte Carlo Masters final for the first time"\}, \{"claim": "Berdych will face either Rafael Nadal or Novak Djokovic"\}, \{"claim": "Berdych will face either Rafael Nadal or Novak Djokovic in the final"\}]

    [\textbf{text}]: 
    
    Tinder only displays the last 34 photos - but users can easily see more. Firm also said it had improved its mutual friends feature.
    
    [\textbf{response}]: 
    
    [\{"claim": "Tinder only displays the last photos"\}, \{"claim": "Tinder only displays the last 34 photos"\}, \{"claim": "Tinder users can easily see more photos"\}, \{"claim": "Tinder said it had improved its feature"\}, \{"claim": "Tinder said it had improved its mutual friends feature"\}]

    Now complete the following:

    [\textbf{text}]: 
    
    \{input\_text\}
    
    [\textbf{response}]:

&
\textbf{[Math Problems]}

You are given a math problem and a potential solution to the math problem. 

\textbf{[Task]}

Your task is to identify all the math calculations that involve arithmetic operations between known real numbers within the potential solution. However, do not include math calculations that contain variable(s).
    
Your response MUST be a list of dictionaries. Each dictionary should contains 2 key - "math\_calculation" and "calculated\_answer", which correspond to the extracted math calculation, and the calculated answer within the potential solution. You MUST only respond in the format as described below. DO NOT RESPOND WITH ANYTHING ELSE. ADDING ANY OTHER EXTRA NOTES THAT VIOLATE THE RESPONSE FORMAT IS BANNED. START YOUR RESPONSE WITH '['.

[\textbf{Response format}]:

[\{"math\_calculation": "Extracted math calculation involving real numbers within the potential solution. Do not include math calculations that contains variable(s). Do not include units such as \$, \%, etc.", "calculated\_answer": "The calculated answer for the extracted math calculation."\},...]

    \textbf{Here are two examples}:

    [\textbf{math problem}]: 
    
    What is the area of a circle with a diameter of 10 inches?

    [\textbf{potential solution}]: 
    
    To find the area, we first calculate the radius as the diameter divided by 2, so the radius is 10/2 = 5 inches. Then, we use the formula for the area of a circle, which is $\pi r^2$. Plugging in the radius we get, Area = $\pi 5^2$ = 78.54 square inches.

    [\textbf{response}]: 
    
    [\{"math\_calculation": "$10 / 2$", "calculated\_answer": "$5$"\}, \{"math\_calculation": "$\pi * 5^2$", "calculated\_answer": "$78.54$"\}]

    [\textbf{math problem}]: 
    
    A store originally sold a shirt for \$45. They are offering a 20\% discount on the shirt. How much will the shirt cost now?

    [\textbf{potential solution}]: 
    
    The discount on the shirt is calculated as 20\% of \$45, which is 0.20 * 45 = \$9. The new price of the shirt after the discount is \$45 - \$9 = \$36.

    [\textbf{response}]: 
    
    [\{"math\_calculation": "0.20 * 45", "calculated\_answer": "9"\}, \{"math\_calculation": "45 - 9","calculated\_answer": "36"\}]

    Now complete the following:

    [\textbf{math problem}]: 
    
    \{input\_question\}

    [\textbf{potential solution}]: 
    
    \{input\_solution\}

    [\textbf{response}]: 

&
\textbf{[Scientific Literature Review]}

You are given a piece of text that mentions some scientific literature. 

[\textbf{Task}]

Your task is to accurately find all papers mentioned in the text and identify the title, author(s), and publication year for each paper.
    The response should be a list of dictionaries, with each dictionary having keys "paper\_title", "paper\_author(s)", and "paper\_pub\_year", which correspond to the title of the paper, the authors of the paper, and the publication year of the paper.
    
    \textbf{The following is the given text}:

    [\textbf{text}]: 
    
    \{input\_text\}

    You MUST only respond in the format as described below. DO NOT RESPOND WITH ANYTHING ELSE. ADDING ANY OTHER EXTRA NOTES THAT VIOLATE THE RESPONSE FORMAT IS BANNED. START YOUR RESPONSE WITH '['.

    [\textbf{Response Format}]:

    [
      \{
        "paper\_title": "Title of the paper.",
        "paper\_author(s)": "Author(s) of the paper.",
        "paper\_pub\_year": "Year of the paper published."
      \},
      ...
    ]
\\
    \bottomrule
    \end{tabular}
    
    \caption{Prompts for Claim Extraction}
    \label{fig:claim_prompt}

\end{figure*}

\begin{figure*}[htbp]
    \scriptsize
    \centering
    \begin{tabular}{@{}p{0.48\textwidth}p{0.48\textwidth}@{}}
\toprule
\textbf{[KB-based QA]}

You are a query generator designed to help users verify a given claim using search engines. Your primary task is to generate a Python list of two effective and skeptical search engine queries. These queries should assist users in critically evaluating the factuality of a provided claim using search engines.
    You should only respond in format as described below (a Python list of queries). PLEASE STRICTLY FOLLOW THE FORMAT. DO NOT RETURN ANYTHING ELSE. START YOUR RESPONSE WITH '['.
    [response format]: ['query1', 'query2']

    Here are 3 examples:
    [claim]: The CEO of twitter is Bill Gates.
    [response]: ["Who is the CEO of twitter?", "CEO Twitter"]

    [claim]: Michael Phelps is the most decorated Olympian of all time.
    [response]: ["Who is the most decorated Olympian of all time?", "Michael Phelps"]

    [claim]: ChatGPT is created by Google.
    [response]: ["Who created ChatGPT?", "ChatGPT"]

    Now complete the following:
    [claim]: {input}
    [response]: 

    ~

    \textbf{[Math Problems]}

    You are given a math calculation and its corresponding calculated answer. 
    
    [\textbf{Task}]
    
    Your task is to write an executable Python snippet that validate the accuracy of the math calculation against the calculated answer. The Python snippet should print 'True' if the calculated answer is correct, and 'False' otherwise.
    
    Your response MUST be a dictionary with key "python\_snippet", which correspond to the executable python snippet.
    
          [math calculation]: \{math\_calculation\}
          
          [calculated answer]: \{calculated\_answer\}
          
          You MUST only respond in the format as described below. DO NOT RESPOND WITH ANYTHING ELSE. ADDING ANY OTHER EXTRA NOTES THAT VIOLATE THE RESPONSE FORMAT IS BANNED. START YOUR RESPONSE WITH '\{'.
    
          [\textbf{Response format}]: 
          
          \{
            "python\_snippet": "An executable Python snippet that validates the accuracy of the math calculation against the calculated answer. The Python snippet should print 'True' if the calculated answer is correct, and 'False' otherwise."
          \}

    &
    \textbf{[Code Potential Solution Generation]}
    
    Please solve the given coding question. Make sure that the solution is optimized and correct. You MUST use Python to solve the coding question.
        Your response MUST be a dictionary with keys "reasoning" and "python\_solution", which correspond to the reasoning and Python implementations of the function \{entry\_point\}.
        The following is the given coding question - 
        [coding question]: \{input\_question\}
        You MUST only respond in the format as described below. DO NOT RESPOND WITH ANYTHING ELSE. ADDING ANY OTHER EXTRA NOTES THAT VIOLATE THE RESPONSE FORMAT IS BANNED. START YOUR RESPONSE WITH '\{'.
        [response format]:
        \{
          "reasoning": "Reasoning for solution.",  
          "python\_solution": "Python implementation of the function \{entry\_point\}. Include only the implementation of the function itself. Ensure the output of the function aligns with its specified return type."
        \}
        
~

\textbf{[Code Unit test Generation]}

Please generate 3 distinct function calls for the given coding question to test the functionality of the function \{entry\_point\} that attempts to solve the provided coding question.

    Your response must be a dictionary with 3 keys - "function\_call\_1", "function\_call\_2", "function\_call\_3", which correspond to the 3 distinct function calls for function \{entry\_point\}.
    The following is the given coding question -

    [coding question]: \{input\_question\}

    You MUST only respond in the format as described below. DO NOT RESPOND WITH ANYTHING ELSE. ADDING ANY OTHER EXTRA NOTES THAT VIOLATE THE RESPONSE FORMAT IS BANNED. START YOUR RESPONSE WITH '\{'.
    
    [response format]: 
    \{
      "function\_call\_1": "First function call for function \{entry\_point\}. Do not include anything else.",
      "function\_call\_2": "Second function call for function \{entry\_point\}. Do not include anything else.",
      "function\_call\_3": "Third function call for function \{entry\_point\}. Do not include anything else."
    \}

\\
    \bottomrule
    \end{tabular}
    \caption{Prompts for Query Generation}
    \label{fig:query_prompt}

\end{figure*}

\begin{figure*}[htbp]
    \scriptsize
        \centering
        \begin{tabular}{@{}p{0.48\textwidth}p{0.48\textwidth}@{}}
    \toprule
\textbf{[KB-based QA]} 

You are given a piece of text. Your task is to identify whether there are any factual errors within the text.
    When you are judging the factuality of the given text, you could reference the provided evidences if needed. The provided evidences may be helpful. Some evidences may contradict to each other. You must be careful when using the evidences to judge the factuality of the given text.
    When 
    The response should be a dictionary with four keys - "reasoning", "factuality", "error", and "correction", which correspond to the reasoning, whether the given text is factual or not (Boolean - True or False), the factual error present in the text, and the corrected text.
    The following is the given text
    [text]: {claim}
    The following is the provided evidences
    [evidences]: {evidence}
    You should only respond in format as described below. DO NOT RETURN ANYTHING ELSE. START YOUR RESPONSE WITH '\{\{'.
    [response format]: 
    \{\{
      "reasoning": "Why is the given text factual or non-factual? Be careful when you said something is non-factual. When you said something is non-factual, you must provide mulitple evidences to support your decision.",
      "error": "None if the text is factual; otherwise, describe the error.",
      "correction": "The corrected text if there is an error.",
      "factuality": True if the given text is factual, False otherwise.
    \}\}

&
\textbf{[Scientific Literature Review]}

Please generate 3 distinct function calls for the given coding question to test the You are provided with two inputs, a string (string1) containing several names, and a list (list1) also containing names. Your task is to assess whether all the last names mentioned in string1 are included in list1.
    
    You should only respond in format as described below. DO NOT RETURN ANYTHING ELSE. START YOUR RESPONSE WITH '\{\{'.
    [response format]: 
    \{\{
      "reasoning": "Explanation on whether all the last names in string1 are found within list1",
      "factuality": This will be True if all last names from string1 are present in list1, and False otherwise.
    \}\}

    Example 1: 
    [string1]: "J. Devlin and M. Chang"
    [list1]: ["Devlin", "M Chang", "Kristina Toutanova"]
    [response]: {{"reasoning": "string1 contains 2 last names 'Devlin' and 'Chang'. Both of these last names are present in list1.", "factuality": True}}

    Example 2: 
    [string1]: "Tom Brown et. al"
    [list1]: ["Y. Lecun", "G. Hinton"]
    [response]: {{"reasoning": "string 1 contains 1 last name 'Brown'. Brown is not present in list1.", "factuality": False}}

    Complete the following:
    [string1]: {string1}
    [list1]: {list2}
    [response]: 

\\
    \bottomrule
    \end{tabular}
    \caption{Prompts for Agreement Verification}
    \label{fig:agreement_prompt}
\end{figure*}

\cleardoublepage

\section{Example cases of \toolname}
We list the example cases of \toolname in each scenario.

\label{sec:appendix:b}

\begin{figure*}[htbp]
     \scriptsize
     \centering
 \begin{tabular}{@{}p{\textwidth}@{}}
 \toprule
 
 \{ 
 
 "claim": "The modern-day version of fortune cookies was invented in the United States.", 
 
 "label": true, 
 "with\_tool\_classification": false, 
 
 "with\_tool\_reasoning": "The given text is not entirely factual. The modern-day version of fortune cookies was not invented in the United States. Most people nowadays believe that fortune cookies were created by a Japanese man named Makoto Hagiwara in 1914 in San Francisco. Hagiwara owned what is now called the Golden Gate Park Japanese Tea Garden, where he served tea and fortune cookies. This is supported by the provided evidences.", 
 
 "queries": ["Where were fortune cookies invented?", "History of fortune cookies in the United States"], 
 
 "evidences": ["Most people nowadays believe that fortune cookies were created by a Japanese man named Makoto Hagiwara in 1914 in San Francisco. Hagiwara owned what is now called the Golden Gate Park Japanese Tea Garden, where he served tea and fortune cookies. Nov 19, 2019", "Most people nowadays believe that fortune cookies were created by a Japanese man named Makoto Hagiwara in 1914 in San Francisco. Hagiwara owned what is now called the Golden Gate Park Japanese Tea Garden, where he served tea and fortune cookies. Nov 19, 2019"] 
 
 \}

 \{ 
 
 "claim": "Sonia Sotomayor is a female US Supreme Court justice", 
 
 "label": true, 
 
 "with\_tool\_classification": false, 
 
 "with\_tool\_reasoning": "The given text is partially factual. While Sonia Sotomayor is indeed a female US Supreme Court justice, the text fails to mention that she was also the first Latina to serve on the Supreme Court. This information is provided in both of the provided evidences.", 
 
 "queries": ["Is Sonia Sotomayor a US Supreme Court justice?", "Sonia Sotomayor gender US Supreme Court"], 
 
 "evidences": ["In 2009, Sonia Sotomayor broke boundaries by becoming the first Latina, and the third woman, to serve on the Supreme Court of the United States, the highest court in the country. Sotomayor was born in New York City to Puerto Rican parents.", "Supreme Court of the United States. Sonia Sotomayor \u2013 the fearless federal trial court judge who saved Major League Baseball from a ruinous 1995 strike \u2013 entered the record book as the first Hispanic and the third woman to serve on the High Court."] 
 
 \}
 \\
 \bottomrule
 \end{tabular}
     \caption{Reasoning errors of \toolname powered by ChatGPT on KB-based QA}
     \label{fig:eval_prompt_single}
 \end{figure*}

\begin{figure*}[htbp]
    
    \scriptsize
    \centering
    \begin{tabular}{@{}p{0.48\textwidth}p{0.48\textwidth}@{}}
\toprule

\textbf{[Fact-checking recent events, discoveries, or news]}

\{

"claim": "Argentina has not won the World Cup since 1986", 

"label": false, 

"with\_tool\_classification": false, 

"with\_tool\_reasoning": "The given text states that Argentina has not won the World Cup since 1986. However, multiple pieces of evidence suggest that Argentina won the World Cup in 2022.", 

"queries": ["Argentina World Cup wins since 1986", "Last time Argentina won World Cup"], 

"evidences": ["Argentina is one of the most successful teams in the tournament's history, having won three World Cups: in 1978, 1986, 2022. Argentina has also been runner up three times: in 1930, 1990 and 2014. In 18 World Cup tournaments, Argentina has 47 victories in 88 matches.", "Argentina is one of the most successful teams in the tournament's history, having won three World Cups: in 1978, 1986, 2022. Argentina has also been runner up ...", "Previously, the last time Argentina won the World Cup was 1986, when it defeated Germany to win its second title in three tournaments.", "Argentina have won the FIFA World Cup three times \u2013 in 1978, 1986 and 2022. They have reached the final six times, including a second-place ...", "The Argentines have won two FIFA World Cups, in 1978 and 1986. They have placed second three times before, in 1930, 1990, and 2014.", "Argentina last won the FIFA World Cup in 1986, defeating West Germany in the championship round, 3-2. Argentina World Cup Wins. Argentina's ..."]

\}

\{

"claim": "The most valuable NFT ever sold is a digital artwork called 'Everydays: The First 5000 Days' ", 

"label": false, 

"with\_tool\_classification": false, 

"with\_tool\_reasoning": "The given text states that the most valuable NFT ever sold is 'Everydays: The First 5000 Days'. According to the provided evidence, 'The First 5000 Days' is the second most expensive NFT ever sold, with 'Merge' being the most expensive one.", 

"queries": ["Most valuable NFT ever sold", "Everydays: The First 5000 Days"], 

"evidences": ["The Most Expensive NFTs Ever Sold  
1. Merge - 2014 US \$91.8M. Pak - Merge. \ldots  
2. The First 5000 Days - 2014 US\$69.3M. Beeple - 2014 The First 5000 Days. \ldots  
3. Clock - 2014 US\$52.7M. Julian Assange \& Pak - 2014 Clock. \ldots  
4. HUMAN ONE - 2014 ~US\$29M. Beeple - 2014 HUMAN ONE. \ldots  
5. CryptoPunk \#5822 - 2014 US\$23.7M. CryptoPunk \#5822.", 
"Everydays: the First 5000 Days: Work of art.", "Everydays: the First 5000 Days is a digital work of art created by Mike Winkelmann, known professionally as Beeple. The work is a collage of 5000 digital images created by Winkelmann for his Everydays series.", "Everydays: the First 5000 Days Artist: Mike Winkelmann.", "Everydays: the First 5000 Days Owner: Vignesh Sundaresan (MetaKovan).", "Everydays: the First 5000 Days Medium: Digital (JPEG)."]

\}

&
\textbf{[Fact-checking high-precision statistics]}

\{

"claim": "Ireland has an obesity rate of 26.9\%", 

"label": false,

"with\_tool\_classification": false, 

"with\_tool\_reasoning": "The given text states that Ireland has an obesity rate of 26.9\%, but the provided evidences show different numbers. The first evidence doesn't provide a clear percentage for obesity, but the second evidence states that the prevalence of obesity in Irish adults is currently 18\%, with men at 20\% and women at 16\%. This contradicts the given text.", 

"queries": ["Ireland obesity rate statistics", "Current obesity rate in Ireland"], 

"evidences": ["Just under four in ten (37\%) of people have a normal weight, six out of ten (37\% overweight and a further 23\% obese) overweight or obese. A smaller difference exists between men and women aged under 25 than those older than this.", "The prevalence of obesity in Irish adults is currently 18\%, with men at 20\% and women at 16\%. A further 47\% of men and 33\% of women are overweight (BMI 25.0-29.9 kg/m2). Since 1990, obesity has more than doubled in men from 8\% to 20\%, and increased from 13\% to 16\% in women."]

\}

\{

"claim": " 'Everydays: The First 5000 Days'  sold for \$69 million", 

"label": false, 

"with\_tool\_classification": false, 

"with\_tool\_reasoning": "The given text is almost factual, but the exact amount for which the artwork was sold is slightly off. Multiple evidences mention the correct amount.", 

"queries": ["Everydays: The First 5000 Days sale price", "
\$69 million Everydays artwork"], 

"evidences": ["The work is a collage of 5000 digital images created by Winkelmann for his Everydays series. Its associated non-fungible token (NFT) was sold for \$69.3 million at Christie's in 2021, making it the most expensive non-fungible token ever.", "\u201cNow there is a way to collect digital art.\u201d The piece that was sold, Everydays: The First 5000 Days, is a collage of Winkelmann's work starting ...", "That day, Winkelmann, a digital artist more commonly known by the handle Beeple, sold \u201cEverydays: The First 5000 Days,\u201d an NFT-linked ...", "\u201cEverydays \u2014 The First 5000 Days,\u201d by the artist known as Beeple, set a record for a digital artwork in a sale at Christie's.", "Artist Beeple, a North Fond du Lac native, sold a digital artwork for \$69 million. Now you can see the same art in a book for a lot less.", "Beeple, aka artist Mike Winkelmann, appeared on \"The ..."]

\}

\\
    \bottomrule
    \end{tabular}
    \caption{Examples of KB-QA claims verified by \toolname powered by GPT-4. The queries generated, evidences collected, the reasoning process, and the final decision of \toolname powered by GPT-4 is reported.}
    \label{fig:KB-QA_example_1}
    \end{figure*}

\begin{figure*}[htbp]
     \scriptsize
     \centering
 \begin{tabular}{@{}p{0.48\textwidth}p{0.48\textwidth}@{}}
 \toprule

\{

"task\_id": "HumanEval/2", 

"prompt": 
\begin{lstlisting}[language=Python]
def truncate_number(number: float) -> float:
    """Given a positive floating point number, 
    it can be decomposed into and integer part 
    (largest integer smaller than given number) 
    and decimals 
    (leftover part always smaller than 1). 
    Return the decimal part of the number."""
\end{lstlisting}

"entry\_point": "truncate\_number", 

"completion": 
\begin{lstlisting}[language=Python]
def truncate_number(number: float) -> float:
    integer_part = number // 1
    decimal_part = number - integer_part
    return decimal_part
\end{lstlisting}

"result": "passed", 

"passed": true, 

"with\_tool\_classification": true, 

"testcases\_input": ["truncate\_number(4.56)", "truncate\_number(0.123)", "truncate\_number(19.999)"], 

"multi\_solutions": [

\begin{lstlisting}[language=Python]
def truncate_number(number: float) -> float:
    return number - int(number)
\end{lstlisting}

\begin{lstlisting}[language=Python]
def truncate_number(number: float) -> float:
    return number - int(number)
\end{lstlisting}

\begin{lstlisting}[language=Python]
def truncate_number(number: float) -> float:
    return number - int(number)
\end{lstlisting}

\begin{lstlisting}[language=Python]
def truncate_number(number: float) -> float:
    integer_part = number // 1
    decimal_part = number - integer_part
    return decimal_part
\end{lstlisting}

]

"exec\_result": [[0.5599999999999996, 0.5599999999999996, 0.5599999999999996, 0.5599999999999996], [0.123, 0.123, 0.123, 0.123], [0.9989999999999988, 0.9989999999999988, 0.9989999999999988, 0.9989999999999988]]

\}

&
\{

"task\_id": "HumanEval/57", 

"prompt": 
\begin{lstlisting}[language=Python]
def monotonic(l: list):
    """Return True if list elements are monotonically 
    increasing or decreasing."""
\end{lstlisting}

"entry\_point": "monotonic", 

"completion": 
\begin{lstlisting}[language=Python]
def monotonic(l: list):
    if len(l) <= 1:
        return True
    increasing = all(l[i] <= l[i+1] for i in range(len(l)-1))
    decreasing = all(l[i] >= l[i+1] for i in range(len(l)-1))
    return increasing or decreasing
\end{lstlisting}

"result": "passed", 

"passed": true, 

"with\_tool\_classification": true, 

"testcases\_input": ["monotonic([1, 2, 3, 4, 5])", "monotonic([5, 4, 3, 2, 1])", "monotonic([1, 3, 5, 4, 7])"],

"multi\_solutions": [ 

\begin{lstlisting}[language=Python]
def monotonic(l: list):
    if len(l) < 2:
        return True
    increasing = decreasing = True
    for i in range(len(l) - 1):
        if l[i] < l[i + 1]:
            decreasing = False
        if l[i] > l[i + 1]:
            increasing = False
        if not increasing and not decreasing:
            return False
    return True
\end{lstlisting}

\begin{lstlisting}[language=Python]
def monotonic(l: list):
    if len(l) < 2:
        return True
    direction = l[1] - l[0]
    for i in range(2, len(l)):
        if direction == 0:
            direction = l[i] - l[i-1]
            continue
        if (l[i] - l[i-1]) * direction < 0:
            return False
    return True
\end{lstlisting}

\begin{lstlisting}[language=Python]
def monotonic(l: list):
    if len(l) < 2:
        return True
    direction = l[1] - l[0]
    for i in range(2, len(l)):
        if direction == 0:
            direction = l[i] - l[i - 1]
            continue
        if (l[i] - l[i - 1]) * direction < 0:
            return False
    return True
\end{lstlisting}

\begin{lstlisting}[language=Python]
def monotonic(l: list):
    if len(l) <= 1:
        return True
    increasing = all(l[i] <= l[i+1] for i in range(len(l)-1))
    decreasing = all(l[i] >= l[i+1] for i in range(len(l)-1))
    return increasing or decreasing
\end{lstlisting}

]

"exec\_result": [[true, true, true, true], [true, true, true, true], [false, false, false, false]]

\}

\\
    \bottomrule
    \end{tabular}
    \caption{Examples of code claims verified by \toolname powered by GPT-4. The queries generated, evidences collected, the reasoning process, and the final decision of \toolname powered by GPT-4 is reported.}
    \label{fig:code_example_1}

\end{figure*}

\begin{figure*}[htbp]
     \scriptsize
     \centering
 \begin{tabular}{@{}p{0.48\textwidth}p{0.48\textwidth}@{}}
 \toprule

\textbf{[ChatGPT]}

\{

"task\_id": "HumanEval/36", 

"prompt": 
\begin{lstlisting}[language=Python]
def fizz_buzz(n: int): 
    """Return the number of times the digit 7 
    appears in integers less than n 
    which are divisible by 11 or 13."""
\end{lstlisting}, 

"entry\_point": "fizz\_buzz", 

"completion": 
\begin{lstlisting}[language=Python]
def fizz_buzz(n: int):
    count = 0
    for i in range(n):
        if i % 11 == 0 or i % 13 == 0:
            if str(i).count('7') > 0:
                count += 1
    return count
\end{lstlisting}

"result": "failed: AssertionError: ", 

"passed": false, 

"testcases\_input": ["fizz\_buzz(50)", "fizz\_buzz(100)", "fizz\_buzz(200)"], 

"multi\_solutions": [
\begin{lstlisting}[language=Python]
def fizz_buzz(n: int):
    count = 0
    for i in range(n):
        if i % 11 == 0 or i % 13 == 0:
            if '7' in str(i):
                count += 1
    return count 
    
def fizz_buzz(n: int):
    count = 0
    for i in range(n):
        if i % 11 == 0 or i % 13 == 0:
            if '7' in str(i):
                count += 1
    return count
    
def fizz_buzz(n: int):
    count = 0
    for i in range(n):
        if i % 11 == 0 or i % 13 == 0:
            count += str(i).count('7')
    return count 
    
def fizz_buzz(n: int):
    count = 0
    for i in range(n):
        if i % 11 == 0 or i % 13 == 0:
            if str(i).count('7') > 0:
                count += 1
    return count
    \end{lstlisting}
]

"exec\_result": [[0, 0, 0, 0], [2, 2, 3, 2], [5, 5, 6, 5]], 

"with\_tool\_classification": true

\}

&
\textbf{[GPT-4]}

\{

"task\_id": "HumanEval/36", 

"prompt": 
\begin{lstlisting}[language=Python]
def fizz_buzz(n: int): 
    """Return the number of times the digit 7 
    appears in integers less than n 
    which are divisible by 11 or 13."""
\end{lstlisting}, 

"entry\_point": "fizz\_buzz", 

"completion": 
\begin{lstlisting}[language=Python]
def fizz_buzz(n: int):
    count = 0
    for i in range(n):
        if i % 11 == 0 or i % 13 == 0:
            if str(i).count('7') > 0:
                count += 1
    return count
\end{lstlisting}

"result": "failed: AssertionError: ",

"passed": false, 

"with\_tool\_classification": false, 

"testcases\_input": ["fizz\_buzz(50)", "fizz\_buzz(100)", "fizz\_buzz(150)"], 

"multi\_solutions": [
\begin{lstlisting}[language=Python]
def fizz_buzz(n: int):
    count = 0
    for i in range(1, n):
        if i % 11 == 0 or i % 13 == 0:
            count += str(i).count('7')
    return count
    
def fizz_buzz(n: int):
    count = 0
    for i in range(1, n):
        if i % 11 == 0 or i % 13 == 0:
            count += str(i).count('7')
    return count
    
def fizz_buzz(n: int):
    count = 0
    for i in range(1, n):
        if i % 11 == 0 or i % 13 == 0:
            count += str(i).count('7')
    return count
    
def fizz_buzz(n: int):
    count = 0
    for i in range(n):
        if i % 11 == 0 or i % 13 == 0:
            if str(i).count('7') > 0:
                count += 1
    return count
]
\end{lstlisting}

"exec\_result": [[0, 0, 0, 0], [3, 3, 3, 2], [4, 4, 4, 3]]

\}

\\
    \bottomrule
    \end{tabular}
    \caption{Examples of code claims verified by \toolname powered by ChatGPT and \toolname powered by GPT-4. The queries generated, evidences collected, the reasoning process, and the final decisions of \toolname powered by ChatGPT and \toolname powered by GPT-4 are reported.}
    \label{fig:code_example_2}

\end{figure*}

\begin{figure*}[htbp]
     \scriptsize
     \centering
 \begin{tabular}{@{}p{0.48\linewidth}p{0.48\linewidth}@{}}
 \toprule

\textbf{[ChatGPT]}

\{

"math\_calculation": "23 * 4319216",

"calculated\_answer": "99305768", 

"label": false, 

"query": {"python\_snippet": "print(23 * 4319216 == 99305768)"}, 

"exec\_result": "False", 

"with\_tool\_classification": false

\}

\{

"math\_calculation": "20/100 x \$10884297.00",

"calculated\_answer": "2176859.40", 

"label": true, 

"query": \{"python\_snippet": "print(abs((20/100) * 10884297.00 - 2176859.40) < 0.01)"\}, 

"exec\_result": "True", 

"with\_tool\_classification": true

\}

&
\textbf{[GPT-4]}

\{

"math\_calculation": "23 * 4319216", 

"calculated\_answer": "99305768", 

"label": false, 

"query": \{"python\_snippet": "calculation = 23 * 4319216; answer = 99305768; print(calculation == answer)"\}, 

"exec\_result": "False", 

"with\_tool\_classification": false

\}

\{

"math\_calculation": "20/100 x \$10884297.00", 

"calculated\_answer": "2176859.40", 

"label": true, 

"query": \{"python\_snippet": "calculation = (20/100) * 10884297.00; answer = 2176859.40; print(abs(calculation - answer) < 0.01)"\}, 

"exec\_result": "True",

"with\_tool\_classification": true

\}

\\
\bottomrule
\end{tabular}
    \caption{Examples of math claims verified by \toolname powered by ChatGPT and GPT-4. The full response of \toolname powered by GPT-4 is reported, including the queries generated, evidences collected, the reasoning process, and the final decision.}
    \label{fig:math_example}
\end{figure*}

\begin{figure*}[htbp]
     \scriptsize
     \centering
 \begin{tabular}{@{}p{0.48\linewidth}p{0.48\linewidth}@{}}
 \toprule

\textbf{[ChatGPT]}

\{

"paper\_title": "BERT: Pre-training of Deep Bidirectional Transformers for Language Understanding", 

"paper\_author(s)": "Jacob Devlin, Ming-Wei Chang, Kenton Lee, Kristina Toutanova", 

"paper\_pub\_year": "2018", 

"label": true, 

"with\_tool\_classification": true, 

"error": []

\}

\{

"paper\_title": "The Impact of Artificial Intelligence on Employment", 

"paper\_author(s)": "Acemoglu and Restrepo", 

"paper\_pub\_year": "2019", 

"label": false, 

"with\_tool\_classification": false, 

"error": ["wrong\_paper\_author(s)", "wrong\_paper\_pub\_year"]

\}

&
\textbf{[GPT-4]}

\{

"paper\_title": "BERT: Pre-training of Deep Bidirectional Transformers for Language Understanding", 

"paper\_author(s)": "Jacob Devlin, Ming-Wei Chang, Kenton Lee, Kristina Toutanova", 

"paper\_pub\_year": "2018", 

"label": true, 

"with\_tool\_classification": true, 

"error": []

\}

\{

"paper\_title": "The Impact of Artificial Intelligence on Employment",

"paper\_author(s)": "Acemoglu and Restrepo", 

"paper\_pub\_year": "2019", 

"label": false, 

"with\_tool\_classification": false, 

"error": ["wrong\_paper\_author(s)", "wrong\_paper\_pub\_year"]

\}

\\
\bottomrule
\end{tabular}
    \caption{Examples of scientific claims verified by \toolname powered by ChatGPT and GPT-4. The full response of \toolname powered by ChatGPT and GPT-4 is reported, including the queries generated, evidences collected, the reasoning process, and the final decision.}
    \label{fig:scientific_example}
\end{figure*}

\begin{figure*}[htbp]
     \scriptsize
     \centering
 \begin{tabular}{@{}p{0.33\linewidth}p{0.33\linewidth}p{0.33\linewidth}@{}}
 \toprule

[\textbf{Reasoning Error}]

\{

"claim": "Jupiter is less dense than Saturn", 

"label": false, 

"with\_tool\_classification": true, 

"with\_tool\_reasoning": "The given text is factual because both provided evidences support the claim that Jupiter is less dense than Saturn. The first evidence provides numerical data on the densities of Jupiter and Saturn, and the second evidence confirms that Saturn is the least dense planet in the solar system.", 

"queries": ["Jupiter vs Saturn density", "Which planet is less dense Jupiter or Saturn?"], 

"evidences": ["The density of Jupiter is 1,33 g/cm\u00b3 whereas the density of Saturn is only 0,687 g/cm\u00b3. If Saturn were the same density as Jupiter, it would have a radius of 30.079 km.", "Jupiter is the largest planet in the solar system, but it's Saturn\u2014the solar system's second largest planet\u2014that takes the prize for least dense. Aug 1, 2019"]

\}

&
[\textbf{Conflicting Evidences}]

\{

"claim": "Jupiter has a density of 1.33 grams per cubic centimeter", 

"label": true,

"with\_tool\_classification": false, 

"with\_tool\_reasoning": "The given text is mostly in line with the provided evidences. There is a minor discrepancy between the two evidences, with one stating the density of Jupiter as 1.326 g/cm\u00b3 and the other as 1.33 g/cm\u00b3. However, the difference is minimal and could be due to rounding.", 

"queries": ["Jupiter density grams per cubic centimeter", "Density of Jupiter"], 

"evidences": ["Density of Jupiter: However, at 1.326 g/cm3, it is also the second-most dense of the gas giants. Feb 17, 2016", "1.33 g/cm\u00b3"]

\}

&
[\textbf{Ambiguity in descriptions}]

\{

"claim": "Fortune cookies are enjoyed by people all over the world.", 

"label": false, 

"with\_tool\_classification": true, 

"with\_tool\_reasoning": "The given text states that fortune cookies are enjoyed by people all over the world. While the evidence provided does not explicitly mention the global enjoyment of fortune cookies, it does mention that 3 billion fortune cookies are made each year globally, with the majority consumed in the US. This suggests that fortune cookies are indeed enjoyed by people in various parts of the world.", 

"queries": ["Origin of fortune cookies", "Popularity of fortune cookies worldwide"], 

"evidences": ["Japan", "There are approximately 3 billion fortune cookies made each year globally, the majority of them consumed in the US. The largest manufacturer of the cookies is Wonton Food, Inc., headquartered in Brooklyn, New York. They make over 4.5 million fortune cookies per day."]

\}
\\
\bottomrule
\end{tabular}
    \caption{Some error cases of \toolname on KB-QA. The full response of \toolname powered by GPT-4 is reported, including the queries generated, evidences collected, the reasoning process, and the final decision is also reported.}
    \label{fig:kbqa_error}
\end{figure*}

\begin{figure*}[htbp]
    \scriptsize
    \centering
\begin{tabular}{@{}p{0.48\linewidth}p{0.48\linewidth}@{}}
\toprule

[\textbf{Limited variety in synthetic testcases}]

\{

"task\_id": "HumanEval/64", 

"prompt": "Write a function \texttt{vowels\_count(s)} which takes a string representing a word as input and returns the number of vowels in the string. Vowels in this case are 'a', 'e', 'i', 'o', 'u'. Here, 'y' is also a vowel, but only when it is at the end of the given word.",

"entry\_point": "vowels\_count", 

"completion": 
\begin{lstlisting}[language=Python]
    def vowels_count(s):
        vowels = ['a', 'e', 'i', 'o', 'u']
        count = 0
        for i in range(len(s)):
            if s[i] in vowels:
                count += 1
            if i == len(s) - 1 and s[i] == 'y':
                count += 1
        return count
\end{lstlisting}

"result": "failed: AssertionError: Test 2", 

"passed": false, 

"with\_tool\_classification": true, 

"testcases\_input": ["vowels\_count(’encyclopedia’)",
"vowels\_count(’rhythm’)", "vowels\_count(’beautiful’)"], 

"multi\_solutions": [ 
    \begin{lstlisting}[language=Python]
    def vowels_count(s):
        vowels = 'aeiou'
        count = 0
        for i, c in enumerate(s):
            if c.lower() in vowels:
                count += 1
            elif c.lower() == 'y' and i == len(s) - 1:
                count += 1
        return count

    def vowels_count(s):
        vowels = 'aeiou'
        count = 0
        for char in s:
            if char.lower() in vowels:
                count += 1
        if s[-1].lower() == 'y':
            count += 1
        return count

    def vowels_count(s):
        vowels = 'aeiou'
        count = 0
        for char in s:
            if char.lower() in vowels:
                count += 1
        if s[-1].lower() == 'y':
            count += 1
        return count
 
    def vowels_count(s):
        vowels = ['a', 'e', 'i', 'o', 'u']
        count = 0
        for i in range(len(s)):
            if s[i] in vowels:
                count += 1
            if i == len(s) - 1 and s[i] == 'y':
                count += 1
        return count
\end{lstlisting}
]

"exec\_result": [[5, 5, 5, 5], [0, 0, 0, 0], [5, 5, 5, 5]]

\}
  
&
\textbf{[Potential errors in code generation]}

\{

"task\_id": "HumanEval/79", 

"prompt": 
\begin{lstlisting}[language=Python]
def decimal_to_binary(decimal): 
"""You will be given a number in decimal form 
and your task is to convert it to binary format. 
The function should return a string, 
with each character representing a binary number. 
Each character in the string will be '0' or '1'. 
There will be an extra couple of 
characters 'db' at the beginning 
and at the end of the string. 
The extra characters are there 
to help with the format. """ 
\end{lstlisting}, 

"entry\_point": "decimal\_to\_binary", 

"completion": 
\begin{lstlisting}[language=Python]
def decimal_to_binary(decimal):
    binary_list = []
    while decimal > 0:
        remainder = decimal % 2
        binary_list.append(str(remainder))
        decimal = decimal // 2
    binary_list.reverse()
    binary_string = ''.join(binary_list)
    return 'db' + binary_string + 'db'
\end{lstlisting}

"result": "failed: AssertionError: ", 

"passed": false, 

"with\_tool\_classification": true, 

"testcases\_input": ["decimal\_to\_binary(10)", "decimal\_to\_binary(255)", "decimal\_to\_binary(0)"], 

"multi\_solutions": [
    \begin{lstlisting}[language=Python]
    def decimal_to_binary(decimal): 
        binary_num = ''
        while decimal > 0:
            remainder = decimal % 2
            decimal = decimal // 2
            binary_num = str(remainder) + binary_num
        return 'db' + binary_num + 'db'
    
    def decimal_to_binary(decimal):
        binary = ''
        while decimal > 0:
            remainder = decimal % 2
            decimal = decimal // 2
            binary = str(remainder) + binary
        return 'db' + binary + 'db'
    
    def decimal_to_binary(decimal): 
        binary = '' 
        while decimal > 0: 
            remainder = decimal % 2 
            decimal = decimal // 2 
            binary = str(remainder) + binary 
        return 'db' + binary + 'db'
    
    def decimal_to_binary(decimal):
        binary_list = []
        while decimal > 0:
            remainder = decimal % 2
            binary_list.append(str(remainder))
            decimal = decimal // 2
        binary_list.reverse()
        binary_string = ''.join(binary_list)
        return 'db' + binary_string + 'db'
    \end{lstlisting}
], 

"exec\_result": [["db1010db", "db1010db", "db1010db", "db1010db"], ["db11111111db", "db11111111db", "db11111111db", "db11111111db"], ["dbdb", "dbdb", "dbdb", "dbdb"]]

\}

\\
    \bottomrule
    \end{tabular}
    \caption{Some error cases of \toolname on code. The full response of \toolname powered by GPT-4 is reported, including the queries generated, evidences collected, the reasoning process, and the final decision is also reported.}
    \label{fig:code_error}

\end{figure*}

\begin{figure*}[htbp]
     \scriptsize
     \centering
 \begin{tabular}{@{}p{\linewidth}@{}}
 \toprule

\textbf{[Round-off error]}

\{ \\
"math\_calculation": "60444034 / 12", 

"calculated\_answer": "5037002.83", 

"label": true, 

"query": {"python\_snippet": "calculation = 60444034 / 12; answer = 5037002.83; print(abs(calculation - answer) < 1e-9)"}, 

"exec\_result": "False", 

"with\_tool\_classification": false

\}

\textbf{[Reasoning error]}

\{

"input": "Kylar went to the store to buy glasses for his new apartment. One glass costs \$5, but every second glass costs only 60\% of the price. Kylar wants to buy 5364765 glasses. How much does he need to pay for them?", 

"target": 21459061.0, 

"response": "Let's assume that Kylar buys x glasses at the full price of \$5 each. Therefore, he will buy (5364765 - x) glasses at a discounted price of 60\% of \$5, which is \$3. Now, we can set up the equation: \\
5x + 3(5364765 - x) = total cost \\

Simplifying the equation, we get: \\
5x + 16094295 - 3x = total cost \\ 
2x + 16094295 = total cost \\

Therefore, the total cost of all the glasses will be 2x + 16094295 dollars. Now, we just need to find the value of x. We know that Kylar wants to buy 5364765 glasses, so we can substitute this value in the equation: \\
2x + 16094295 = total cost \\
2x + 16094295 = 5,364,765 * 5 \\
2x + 16094295 = 26823825 \\
2x = 10729530 \\
x = 5364765 \\

Therefore, Kylar needs to pay 2x + 16094295 dollars, which is: 2(5364765) + 16094295 = 26823825. \\

Hence, Kylar needs to pay \$26823825 for all the glasses.", 

"answer": "26823825", 

"annotation": false, 

"claims": [\{"math\_calculation": "5364765 - x", "calculated\_answer": "5364765 - x", "label": true\}, \{"math\_calculation": "0.60 * 5", "calculated\_answer": "3", "label": true\}, \{"math\_calculation": "5x", "calculated\_answer": "5x", "label": true\}, \{"math\_calculation": "3(5364765 - x)", "calculated\_answer": "16094295 - 3x", "label": true\}, \{"math\_calculation": "5x + 16094295 - 3x", "calculated\_answer": "16094295 + 2x", "label": true\}, \{"math\_calculation": "2x + 16094295", "calculated\_answer": "total cost", "label": "null"\}, \{"math\_calculation": "5364765 * 5", "calculated\_answer": "26823825", "label": true\}, \{"math\_calculation": "2(5364765)", "calculated\_answer": "10729530", "label": true\}, \{"math\_calculation": "2x", "calculated\_answer": "10729530", "label": "null"\}]

{"math\_calculation": "5364765 - x", "calculated\_answer": "5364765 - x", "label": true, "query": {"python\_snippet": "x = 0; math\_calculation = 5364765 - x; calculated\_answer = 5364765 - x; print(math\_calculation == calculated\_answer)"}, "exec\_result": "True", "with\_tool\_classification": true}
{"math\_calculation": "0.60 * 5", "calculated\_answer": "3", "label": true, "query": {"python\_snippet": "calculation = 0.60 * 5; answer = 3; print(calculation == answer)"}, "exec\_result": "True", "with\_tool\_classification": true}
{"math\_calculation": "5x", "calculated\_answer": "5x", "label": true, "query": {"python\_snippet": "def validate\_calculation(calculation, answer): return calculation == answer; print(validate\_calculation('5x', '5x'))"}, "exec\_result": "True", "with\_tool\_classification": true}
{"math\_calculation": "3(5364765 - x)", "calculated\_answer": "16094295 - 3x", "label": true, "query": {"python\_snippet": "x = 1; result1 = 3 * (5364765 - x); result2 = 16094295 - 3 * x; print(result1 == result2)"}, "exec\_result": "True", "with\_tool\_classification": true}
{"math\_calculation": "5x + 16094295 - 3x", "calculated\_answer": "16094295 + 2x", "label": true, "query": {"python\_snippet": "x = 1; print((5 * x + 16094295 - 3 * x) == (16094295 + 2 * x))"}, "exec\_result": "True", "with\_tool\_classification": true}
{"math\_calculation": "2x + 16094295", "calculated\_answer": "total cost", "label": "null", "query": {"python\_snippet": "x = 5; math\_calculation = 2 * x + 16094295; calculated\_answer = 16094305; print(math\_calculation == calculated\_answer)"}, "exec\_result": "True", "with\_tool\_classification": true}
{"math\_calculation": "5364765 * 5", "calculated\_answer": "26823825", "label": true, "query": {"python\_snippet": "calculation = 5364765 * 5; answer = 26823825; print(calculation == answer)"}, "exec\_result": "True", "with\_tool\_classification": true}
{"math\_calculation": "2(5364765)", "calculated\_answer": "10729530", "label": true, "query": {"python\_snippet": "calculation = 2 * 5364765; answer = 10729530; print(calculation == answer)"}, "exec\_result": "True", "with\_tool\_classification": true}
{"math\_calculation": "2x", "calculated\_answer": "10729530", "label": "null", "query": {"python\_snippet": "x = 5364765; print(2 * x == 10729530)"}, "exec\_result": "True", 
"with\_tool\_classification": true}

\}

\\
\bottomrule
\end{tabular}
    \caption{Some error cases of \toolname on math. The full response of \toolname powered by GPT-4 is reported, including the queries generated, evidences collected, the reasoning process, and the final decision is also reported.}
    \label{fig:math_error}
\end{figure*}

\begin{figure*}[htbp]
     \scriptsize
     \centering
 \begin{tabular}{@{}p{\linewidth}@{}}
 \toprule

\textbf{[Errors in title matching]}

\{

"paper\_title": "MDMA-assisted psychotherapy for treatment of PTSD: study design and rationale for phase 3 trials based on pooled analysis of six phase 2 randomized controlled trials", 

"paper\_author(s)": "Mithoefer et al.", 

"paper\_pub\_year": "2019", 

"label": true, 

"with\_tool\_classification": false, 

"error": ["wrong\_paper\_title"]

\}

\textbf{[Errors in author matching]}

\{

"paper\_title": "Language Models are Unsupervised Multitask Learners", 

"paper\_author(s)": "Alec Radford, Jeffrey Wu, Rewon Child, David Luan, Dario Amodei, Ilya Sutskever", 

"paper\_pub\_year": "2019", 

"label": true, 

"with\_tool\_classification": false, 

"error": ["wrong\_paper\_author(s)"]

\}

\\
\bottomrule
\end{tabular}
\caption{Some error cases of \toolname on scientific. The full response of \toolname powered by GPT-4 is reported, including the queries generated, evidences collected, the reasoning process, and the final decision is also reported.}
    \label{fig:scientific_error}
\end{figure*}

\end{document}